\documentclass[twoside,11pt]{article}

\usepackage[abbrvbib, preprint]{jmlr2e}

\jmlrheading{1}{2000}{1-48}{4/00}{10/00}{meila00a}{Henning Lange, Steve L. Brunton, and J. Nathan Kutz}

\ShortHeadings{Spectral Methods for Long-term Time Series Prediction}{Lange, Brunton, and Kutz}
\firstpageno{1}

\title{From Fourier to Koopman: Spectral Methods for Long-term Time Series Prediction}
\author{\name Henning Lange \email helange@uw.edu \\
       \addr Department of Applied Mathematics\\
       University of Washington\\
       Seattle, WA 98195-4322, USA
       \AND
        \name Steven L. Brunton \email sbrunton@uw.edu \\
       \addr Department of Mechanical Enigineering\\
       University of Washington\\
       Seattle, WA 98195-4322, USA
       \AND
       \name J. Nathan Kutz \email kutz@uw.edu \\
       \addr Department of Applied Mathematics\\
       University of Washington\\
       Seattle, WA 98195-4322, USA}

\editor{Kevin Murphy and Bernhard Sch{\"o}lkopf}
\usepackage{natbib}
\usepackage{graphicx}
\usepackage{amsmath}
\usepackage{amssymb}
\usepackage{hyperref}

\usepackage{multirow}

\usepackage{algorithm}
\usepackage{algpseudocode}

\usepackage{setspace}
\begin{document}

\maketitle

\begin{abstract}
We propose spectral methods for long-term forecasting of temporal signals stemming from linear and nonlinear quasi-periodic dynamical systems. For linear signals, we introduce an algorithm with similarities to the Fourier transform but which does not rely on periodicity assumptions, allowing for forecasting given potentially arbitrary sampling intervals. We then extend this algorithm to handle nonlinearities by leveraging Koopman theory. The resulting algorithm performs a spectral decomposition in a nonlinear, data-dependent basis. The optimization objective for both algorithms is highly non-convex.  However, expressing the objective in the frequency domain allows us to compute global optima of the error surface in a scalable and efficient manner, partially by exploiting the computational properties of the Fast Fourier Transform. Because of their close relation to Bayesian Spectral Analysis, uncertainty quantification metrics are a natural byproduct of the spectral forecasting methods. We extensively benchmark these algorithms against other leading forecasting methods on a range of synthetic experiments as well as in the context of real-world power systems and fluid flows.
 %The efficacy of the algorithms is evaluated on synthetic experiments as well as in the context of predicting signals in the realms of power systems and fluid flows. 
\end{abstract}

\section{Introduction}

Forecasting of time series data is fundamental to almost every discipline in science and engineering, with applications in oceanography, epidemiology, meteorology, economics, etc~\citep{armstrong1985long}.  Indeed, {\em time-series analysis} has been historically a core discipline in statistical sciences where prediction accuracy and uncertainty quantification are critical for characterizing the underlying complex dynamical system measured.  Such future state predictions also play a fundamental role in control theory where improved forecasting can in turn improve the latency time of control laws~\citep{box2015time}.  
While there are a host of effective short-time forecasting methods~\citep{hamilton1994time}, generic mathematical architectures that allow for stable long-term forecasts, while mitigating compounding errors, are lacking.

The fundamental problem of forecasting has generated a diverse set of approaches for producing future state predictions and uncertainty metrics. Most of the pre-existing approaches are of the following type: a parameterized function $f_\Theta: \mathbb{R}^{T' \times N} \rightarrow \mathbb{R}^N$ is introduced and tasked with predicting the next temporal snapshot given a collection of previous ones by minimizing some loss function with respect to the parameters $\Theta$. Prominent members of this family of approaches include Box-Jenkins models~\citep{box2015time,pankratz2009forecasting}, for example ARIMA or SARIMAX, and several neural network approaches, including Recurrent Neural Networks, such as Long Short-Term Memory (LSTM)~\citep{hochreiter1997long}, Gated Recurrent Units~\citep{chung2014empirical} and most Echo State Networks~\citep{jaeger2007echo}, as well as Reservoir Computing approaches~\citep{paquot2012optoelectronic} and hybrid approaches that combine wavelets and Neural Networks~\citep{jiang2005dynamic}. Since $f_\Theta$ usually only allows one to obtain an estimate of the value the signal will take in the next time step, in order to allow for long-term forecasting, estimates must be recursively fed back into $f_\Theta$. The main drawback of approaches that recursively perform one-step-ahead predictions, i.e. an approach that iteratively performs a succession of local decisions, is that errors compound over time. %Since $f_\Theta$ usually only allows one to obtain an estimate of the value the signal will take in the next time step, i.e. $\hat{\mathbf{x}}_{T+1}$, in order to allow for long-term forecasting, estimates $\hat{\mathbf{x}}_{T+i}$ need to be recursively fed back into $f_\Theta$ to obtain $\hat{\mathbf{x}}_{T+i+1}$.

In this work, ideas from operator theory, signal processing, and machine learning are combined to (i) derive generic and, under some conditions, optimal algorithms for long-term forecasting that do not require knowledge of the dynamics and that mitigate compounding errors, and (ii) derive practical error bounds and uncertainty metrics of the resulting algorithms as functions of noise and amounts of data provided. Because the contributions in this work touch multiple disciplines, they can be understood from different perspectives.

From the viewpoint of signal processing, we leverage {\em Bayesian Spectral Analysis}~\citep{bretthorst2013bayesian} and derive an algorithm that performs a Fourier decomposition in an optimal nonlinear oscillatory basis. This nonlinear basis is learned from data and is optimal in the least-squares sense. Because of a symmetry relationship, we can leverage the {\it Fast Fourier Transform} to obtain model parameters in a fast and scalable way.

\begin{figure}[t]
    \centering
    \vspace{-.35in}
    \includegraphics[width=0.75\linewidth]{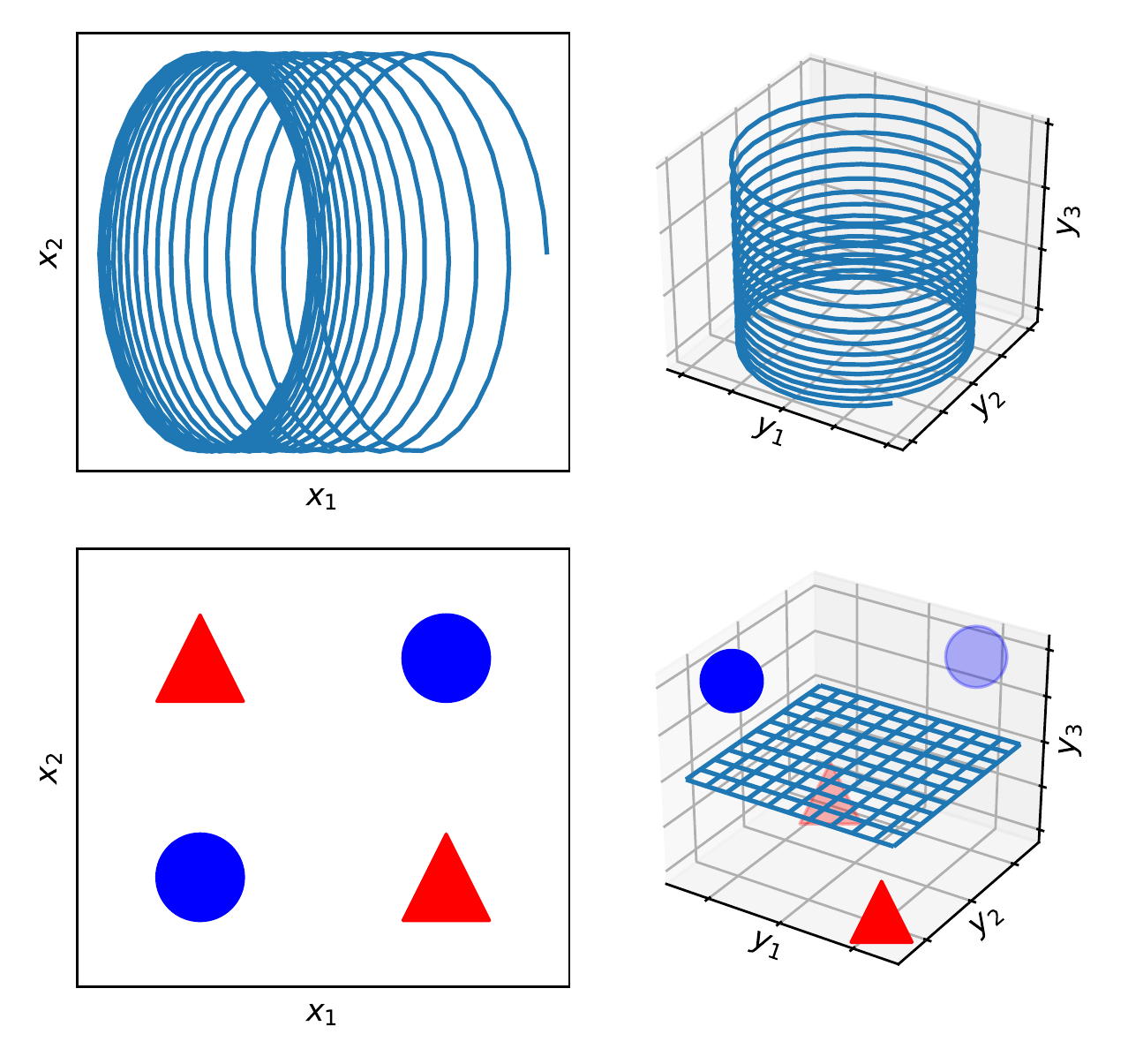}
    \vspace{-.2in}
    \caption{Top: Left shows the phase portrait of a nonlinear system that can be described as a linear system in three dimensions, as shown in the right panel. Bottom: The \emph{exclusive-or} problem does not have a linear solution in 2D, but does in 3D.}
    \label{fig:cover_koop}
    \vspace{-.3in}
\end{figure}

When adopting a Machine Learning perspective, because the theoretical and practical contributions of this work are primarily based on Koopman theory~\citep{mezic2015applications}, they can be understood by analogy to Cover's theorem, which enabled the breakthroughs that have led to the development of Kernel Methods~\citep{scholkopf2001learning} and Deep Learning~\citep{lecun2015deep}. Succinctly, Cover's theorem states that for any data set of finite length there always exists a nonlinear transformation such that in the transformed space, the data can be described by linear methods, i.e. it is {\it linearly separable}. Analogously, Koopman theory postulates that any nonlinear dynamical system can be lifted by a time-invariant but nonlinear function into a space in which its time evolution can be described by linear methods, i.e. by a linear dynamical system~\citep{kutz2016dynamic}. This analogy is depicted graphically in Fig.~\ref{fig:cover_koop}.

For finite amounts of data, Cover's and Koopman's theorem can be made to hold in a trivial way by projecting the data into an $N$ dimensional space, where $N$ denotes the number of data points. For Cover's theorem, a nonlinear function can be devised that maps the $n$-th data point to the indicator vector that is 0 everywhere except for the $n$-th position, where it is 1. By doing so, every finite data set can be made linearly separable by constructing a weight vector ${\bf w} \in \mathcal{R}^N$ that is 1 for all data points belonging to one class and 0 for the other, i.e. by simply enumerating the training set. Similarly, Koopman's theorem can be made true in an $N$ dimensional space by the means of the discrete Fourier transform. Since sine and cosine pairs are linear dynamical systems and every sequence of $N$ equidistantly sampled data points can be explained by $N$ equidistant Fourier modes. However, both trivial solutions are not useful in practical scenarios since they do not allow for generalization. Note that the inability of the trivial solution of Koopman's theorem to generalize stems from implicit periodicity assumptions of the discrete Fourier transform. We will show later how these implicit periodicity assumptions can be broken in order to allow for generalization.

We leverage Koopman theory for forecasting by constructing the linear {\it Koopman operator} that characterizes the time evolution of a dynamical system, even chaotic systems. However in practice, such chaotic systems result in a Koopman operator that is infinite dimensional. In this work, we focus our investigation on systems for which a finite dimensional Koopman operator exists~\cite{Brunton2016plosone}, and furthermore restrict the scope to systems that preserve energy, or persist over long times, i.e. \emph{quasi-periodic} systems. Mathematically speaking, we require the following property of the time series data $x_t$: $$\lim_{T \rightarrow \infty} \mathcal{F} [f^{-1}(x_1), ..., f^{-1}(x_T)] = \sum_{n=1}^N \delta(\omega - k_n)$$
where $\mathcal{F}$ is the Fourier transform, $f$ is a function, and $f^{-1}$ its pre-image%. , and finite $N$. 
When systems fulfill these properties, we show empirically that stable long-term forecasting can be accomplished.

\section{Fourier and Forecasting}
\label{sec:fourier_intro}

When Fourier studied heat transfer in the early 19th century, he discovered that solutions to the heat equation can be simply represented as a superposition of sines and cosines~\citep{fourier1822theorie}. This finding was generalized and ultimately resulted in what is known today as the Fourier Transform. In the 1960s, Cooley and Tukey (re)discovered the Fast Fourier Transform (FFT), which allows for the computation of coefficients of the discrete Fourier transform in $\mathcal{O}(T \log T)$ given a signal sampled at $T$ equidistant points in time~\citep{cooley1965algorithm}. The computational speed up relies on evaluating the Fourier transform at integer multiples of $\frac{1}{T}$, which in turn implicitly assumes that the signal is periodic with period $T$ since 
$$x_{t+T} = \sum_{n=1}^N a_n e^{-j2\pi (t+T) \frac{n}{T}} = x_t$$ 
This assumption, unfortunately, makes the FFT unsuitable for forecasting anything but strictly $T$-periodic signals. However, in the following, we will treat the FFT as a computationally efficient optimization tool that, in conjunction with gradient descent, allows us to fit a linear dynamical system $y_t$ to an observed signal $x_t$. The strategy is as follows: Given an optimization problem, we will derive an expression for a loss as a function of model parameters in the frequency domain and then use the FFT to evaluate the loss and pick optimal parameters. This strategy is similar to the initial idea of Fourier, i.e. to express an otherwise difficult problem in the frequency domain in which finding a solution is significantly easier. As we will show later, the difficulty of finding a solution in time-domain stems from the fact that the optimization objective is highly non-convex. However, note that in order to leverage the computational efficiency of the FFT, implicit periodicity assumption are being made. While these periodicity assumptions bring about computational efficiency, they prohibit forecasting but they can easily be broken by subsequently applying gradient descent to improve an initial guess provided by the FFT.\\

Let $\mathbf{y}_t \in \mathbb{R}^m$ be measurements of a linear dynamical system, i.e. $\mathbf{y}_t = \mathbf{B}\mathbf{y}_{t-1}$. The goal is to infer an optimal linear mapping between a measured signal $\mathbf{x}_t \in \mathbb{R}^n$ and $\mathbf{y}_t$. Specifically, the goal is to solve:
\begin{align*}
    \text{minimize }& E(\mathbf{A},\mathbf{B}) = \sum_{t=1}^T (\mathbf{x}_t - \mathbf{Ay}_t)^2\\
    \text{subject to } &\mathbf{y}_t = \mathbf{By}_{t-1}
\end{align*}

Since we assume $\mathbf{x}_t$ to be quasi-periodic, we in turn assume $\mathbf{B}$ to have strictly imaginary eigenvalues. For the continuous time solution of $\mathbf{y}_t$, the following holds:
\begin{align}
    \mathbf{y}_t &= \sum _{i=1}^{n}\left(\mathbf{l}_{i}\cdot \mathbf{y}_{0}\right)\mathbf{r}_{i}e^{\lambda _{i}t} \label{k0_lti}
\end{align}
where $\mathbf{l}_i$, $\mathbf{r}_i$ and $\lambda_i$ denote the $i$th left and right eigenvector and the $i$th eigenvalue of $\mathbf{B}$, respectively. Note, because eigenvalues and eigenvectors of real-valued square matrices come in conjugate pairs, we can rewrite equation (\ref{k0_lti}) as:
\begin{align*}
    \mathbf{y}_t &= \sum_{i=1}^{n/2} \left(\mathbf{l}_{i}\cdot \mathbf{y}_{0}\right)\mathbf{r}_i e^{\lambda_{i}t} + (\left(\mathbf{l_{i}}\cdot \mathbf{y}_{0}\right)\mathbf{r}_i)^* e^{-\lambda_{i}t} \\
                 &= \sum_{i=1}^{n/2} \mathbf{b}_i e^{\lambda_{i}t} + \mathbf{b}_i^* e^{-\lambda_{i}t}
\intertext{Since we assume $\lambda_i$ to be strictly imaginary, let $\lambda_i = j\omega_i$. We can further simplify:}                
    \mathbf{y}_t &= \sum_{i=1}^{n/2} 2 Re(\mathbf{b}_i)\cos(\omega_i t) + 2 Im(\mathbf{b}_i)\sin(\omega_i t)\\  
                &= 2\begin{bmatrix}
                        Re(\mathbf{b}_1) & \cdots & Re(\mathbf{b}_{n/2}) & Im(\mathbf{b}_1) & \cdots & Im(\mathbf{b}_{n/2})
                    \end{bmatrix}
                     \begin{bmatrix}
                           \cos(\omega_1 t)\\
                           \vdots \\
                           \cos(\omega_{n/2} t)\\
                           \sin(\omega_1 t)\\
                           \vdots \\
                           \sin(\omega_{n/2} t)\\
                    \end{bmatrix}\\
                &= 2\overline{\mathbf{B}} \begin{bmatrix}
       \cos(\Vec{\omega} t)\\
       \sin(\Vec{\omega} t)
     \end{bmatrix}
\end{align*}

Without loss of generality, because $\mathbf{A}$ is a free-parameter of the optimization problem, we can omit $2\overline{\mathbf{B}}$ and absorb this into $\mathbf{A}$:

\begin{align}
    \mathbf{y}_t = \begin{bmatrix}
       \cos(\Vec{\omega} t)\\
       \sin(\Vec{\omega} t)
     \end{bmatrix} := \Omega(\Vec{\omega} t) \label{eq:cos}
\end{align}

Thus, under the assumption that $\mathbf{B}$ only has imaginary eigenvalues, we can rewrite the minimization problem in the following way:

\begin{align}
    L_i(\mathbf{A}, \Vec{\omega}, t) &:= [(\mathbf{x}_t)_i - (\mathbf{A}\Omega(\Vec{\omega} t))_i]^2  \nonumber \\
    L(\mathbf{A}, \Vec{\omega}, t) &=  ||\mathbf{x}_t - \mathbf{A}\Omega(\Vec{\omega} t)||_2^2 &\text{(temporally local loss)} \nonumber \\
                       &=  \sum_{i=1}^n L_i(\mathbf{A}, \Vec{\omega}, t) \nonumber\\
    E(\mathbf{A}, \Vec{\omega}) &= \sum_{t=1}^T L(\mathbf{A}, \Vec{\omega}, t) = \sum_{t=1}^T \sum_{i=1}^n L_i(\mathbf{A}, \Vec{\omega}, t) &\text{(global loss)} \label{eq:mini_fourier}
\end{align}

The question arises which optimization procedure to employ to solve (\ref{eq:mini_fourier}) for $\Vec{\omega}$. Note that the assumption of suitable convexity required for gradient descent to perform well does not hold. This can be understood by realizing that the sensitivity of the loss with respect to $\Vec{\omega}$ grows linearly in time which in turn can be understood easily by investigating the gradients of the loss with respect to $\Vec{\omega}$.

\begin{align*}
    \frac{\partial L(\mathbf{A}, \Vec{\omega}, t)}{\partial \Vec{\omega}} = 2t \begin{bmatrix}
       -\sin(\Vec{\omega} t)\\
       \cos(\Vec{\omega} t) 
     \end{bmatrix}[\mathbf{x}_t - \mathbf{A}\Omega(\Vec{\omega} t)]
\end{align*}

This implies that the magnitude of gradients grows linearly with time which in turn entails that gradient descent disproportionately considers the loss function of temporal snapshots close to $T$. This effect can also be understood by realizing that $L(A, \Vec{\omega}, t)$ is periodic in $\frac{2 \pi}{t}$ with respect to $\Vec{\omega}$ which follows directly from the properties of the sine and cosine functions, since:

\begin{align*}
    L(\mathbf{A}, \Vec{\omega} + \frac{2 \pi}{t}, t) &=  ||\mathbf{x}_t - \mathbf{A}\Omega((\Vec{\omega}+\frac{2 \pi}{t}) t)||_2^2 \\
                       &=  ||\mathbf{x}_t - \mathbf{A}\Omega(\Vec{\omega} t)||_2^2 \\
                       &= L(\mathbf{A}, \Vec{\omega}, t) 
\end{align*}

%\begin{figure}
%    \centering
%    \includegraphics[width=\linewidth]{plots/time_sensitivity.pdf}
%    \caption{This figure shows the reason why the optimization objective is highly non-convex: Temporally local %losses oscillate with period $2\pi / t$}
%    \label{fig:time_sensitivity}
%\end{figure}

Note that this entails that the temporally local loss functions close to $T$ will oscillate rapidly and therefore create local minima in which gradient descent will get stuck.\\

In order to overcome these issues, we will derive an algorithm that under some conditions finds global minimizers by performing coordinate descent on $E(A, \Vec{\omega})$. Coordinate descent updates model parameters in a sequential fashion starting with an initial guess of $\Vec{\omega}^{(0)}$.

\begin{align*}
    \omega_i^{(k)} = \arg \min_{\omega_i}  E(\omega_{1}^{(k)}, .., \omega_{i-1}^{(k)}, \omega_i, \omega_{i+1}^{(k-1)}, .., \omega_m^{(k-1)}) \
\end{align*}

Under some conditions \citep{tseng2001convergence}, for $k=1,2,3,...$, $\Vec{\omega}^{(k)}$ converges to a minimizer of $E$. We will now show that the update-step in the direction of $\omega_i$ can be performed in a computationally efficient manner using the FFT. Let $\mathcal{M}_i = \{A, \omega_{1}, .., \omega_{i-1}, \omega_{i+1}, .., \omega_m\}$, i.e. the set of all model parameters except $\omega_i$. We will derive an analytic expression for $E(\omega_i; \mathcal{M}_i)$ in the frequency domain which allows us, by applying the FFT, to compute the entire error surface and pick the optimal value for $\omega_i$.\\

Let $L_l(\omega_i;\mathcal{M}_i,t)$ be the temporally local loss as a function of only $\omega_i$, i.e. while keeping all other model parameters fixed and only considering the $l$th dimension of the loss and let $I(i) = \{1,..,i-1,i+1,..,i+n/2-1,i+n/2+1,..,n\}$ be the set of all indices sans $i$ and $i+n/2 := i_2$, then:
\begin{align*}
    L_l(\omega_i;\mathcal{M}_i,t) &= [(\mathbf{x}_t)_l - (\mathbf{A}\Omega(\Vec{\omega}t))_l]^2 \\
                                   &= [\underbrace{(\mathbf{x}_t)_l - \sum_{k \in I(i)} \mathbf{A}_{l,k} \Omega(\Vec{\omega}t)_k}_{= R(i)_{t,l}} - \mathbf{A}_{l,i} \cos(\omega_i t) - \mathbf{A}_{l,i_2} \sin(\omega_i t)]^2
\end{align*}

Note that $R(i)$ defines the residual that captures the difference between $Ay_t$ and $x_t$ but ignores all terms that involve $\omega_i$.

\begin{align}
    E_l(\omega_i;\mathcal{M}_i) :=& \sum_t L_l(\omega_i;\mathcal{M}_i,t)\nonumber \\
                    =& \sum_t (\mathbf{R}(i)_{l,t} - \mathbf{A}_{i,l} \cos(\omega_it) - \mathbf{A}_{i_2,l} \sin(\omega_i t))^2\nonumber \\
                    =& \sum_t [\mathbf{R}(i)_{l,t}^2 - \mathbf{A}_{i,l}^2 \cos^2(\omega_i t) - \mathbf{A}_{i_2,l}^2 \sin^2(\omega_i t)\nonumber \\
                    &- 2\mathbf{R}(i)_{l,t}\mathbf{A}_{i,l} \cos(\omega_i t) -2\mathbf{R}(i)_{l,t}\mathbf{A}_{i_2,l} \sin(\omega_i t)\nonumber \\
                    &+ 2\mathbf{A}_{i,l} \mathbf{A}_{i_2,l} \cos(\omega_i t)\sin(\omega_i t)] \label{eq:2AAcossin}
\end{align}

We proceed by investigating this loss function at extrema of $\mathbf{A}$.

\begin{align*}
    \frac{\partial E_l(\omega_i;\mathcal{M}_i)}{\partial  \mathbf{A}_{i,l}} &= \sum_{t} -2\mathbf{A}_{i,l} \cos^2(\omega_it) - 2\mathbf{R}(i)_{l,t} \cos(\omega_it) + 2\mathbf{A}_{i_2,l} \cos(\omega_it)\sin(\omega_it)\\
    \frac{\partial E_l(\omega_i;\mathcal{M}_i)}{\partial  \mathbf{A}_{i_2,l}} &= \sum_{t} -2\mathbf{A}_{i_2,l} \sin^2(\omega_it) - 2\mathbf{R}(i)_{l,t} \sin(\omega_it) + 2\mathbf{A}_{i,l} \cos(\omega_it)\sin(\omega_it)
\end{align*}

From $\frac{\partial E_l(\omega_i;\mathcal{M}_i)}{\partial  \mathbf{A}_{i,l}} = 0$ and $\frac{\partial E_l(\omega_i;\mathcal{M}_i)}{\partial  \mathbf{A}_{i_2,l}} = 0$, we can derive the following respectively:

\begin{align}
    \mathbf{A}_{i_2,l} \sum_t \cos(\omega_it)\sin(\omega_it) &= \sum_{t} \mathbf{A}_{i,l} \cos^2(\omega_it) + \mathbf{R}(i)_{l,t} \cos(\omega_it)\label{eq:A1} \\
    \mathbf{A}_{i,l} \sum_t \cos(\omega_it)\sin(\omega_it) &= \sum_{t} \mathbf{A}_{i_2,l} \sin^2(\omega_it) + \mathbf{R}(i)_{l,t} \sin(\omega_it)\label{eq:A12}
\end{align}

Plugging both of these findings into (\ref{eq:2AAcossin}) yields:

\begin{align*}
    E_l(\omega_i;\mathcal{M}_i) =& \sum_t [\mathbf{R}(i)_{l,t}^2 - \mathbf{R}(i)_{l,t}\mathbf{A}_{i,l} \cos(\omega_i t) - \mathbf{R}(i)_{l,t}\mathbf{A}_{i_2,l} \sin(\omega_i t)]
\end{align*}

Since our goal is to employ the FFT to evaluate $E(\omega_i;\mathcal{M}_i)$, we assume that all $\omega_i = \frac{k}{T}$ with $k\in \mathbb{N}$. This entails that $\sum_t^T \cos^2(\omega_i t) = \sum_t^T \sin^2(\omega_i t) = T/2$, $\sum_t^T \cos(\omega_i t)\sin(\omega_i t) = 0$. We can then deduce from (\ref{eq:A1}) and (\ref{eq:A12}) respectively, that:

\begin{align*}
    \mathbf{A}_{i,l} = \frac{2\mathbf{R}(i)_{l,t} \cos(\omega_i t)}{T}\\
    \mathbf{A}_{i_2,l} = \frac{2\mathbf{R}(i)_{l,t} \sin(\omega_i t)}{T}
\end{align*}
Therefore:
\begin{align*}
    E_l(\omega_i;\mathcal{M}_i) =& \sum_t^T \mathbf{R}(i)_{l,t}^2 - \frac{2}{T}(\sum_t^T \mathbf{R}(i)_{l,t} \cos(\omega_i t))^2 - \frac{2}{T}(\sum_t^T \mathbf{R}(i)_{l,t} \sin(\omega_i t))^2
\end{align*}

Let $\mathbf{\hat{R}}(i)_{l,\omega_i}$ be the Fourier transform of $\mathbf{R}(i)_{l,t}$, then:
\begin{align*}
    E(\omega_i;\mathcal{M}_i) =& \sum_l^n( \sum_t^T \mathbf{R}(i)_{l,t}^2 - \frac{2}{T} Re(\mathbf{\hat{R}}(i)_{l,\omega_i})^2 - \frac{2}{T} Im(\mathbf{\hat{R}}(i)_{l,\omega_i})^2)\\
                         =& ||\mathbf{R}(i)||_F - \sum_l^n \frac{2}{T} |\mathbf{\hat{R}}(i)_{l,\omega_i}|^2
\end{align*}

\begin{figure}
    \centering
    \vspace{-.3in}
    \includegraphics[width=\linewidth]{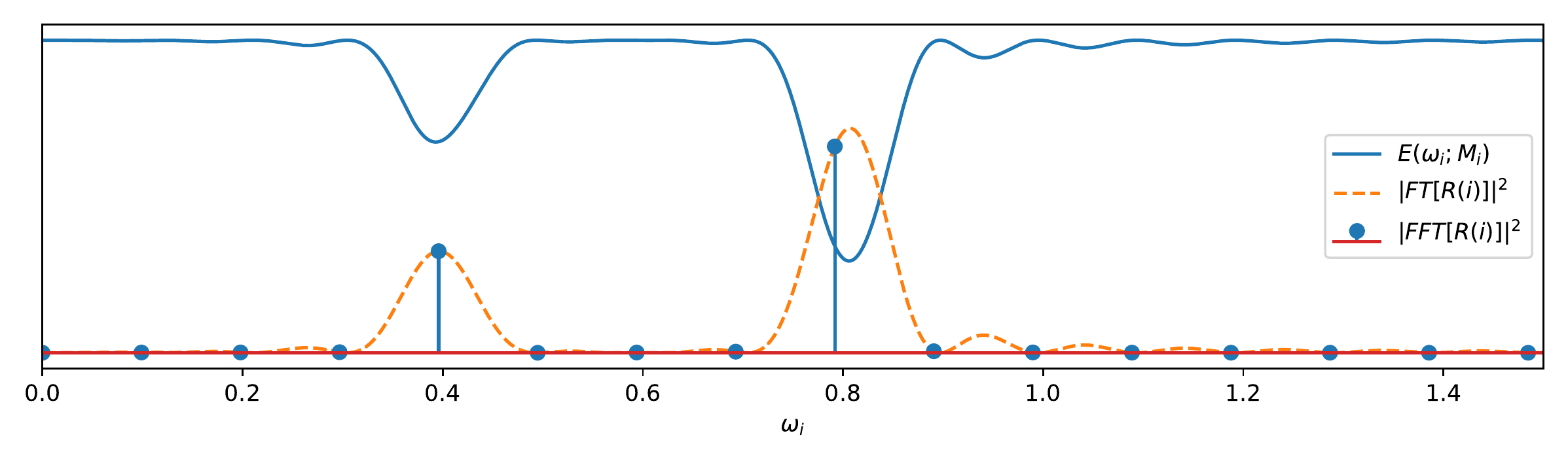}
    \vspace{-.4in}
    \caption{Symmetry between the squared error, i.e. $E(\omega_i;\mathcal{M}_i)$, and the Fourier transform of the residual. Colloquially speaking, the absolute value of the Fourier coefficients moved up and flipped on the x-axis constitutes the error surface.}
    \vspace{-.2in}
    \label{fig:dft_symmetry}
\end{figure}

Thus, there is a symmetry relationship between the error surface and the Fourier coefficients of the residual. This symmetry is sketched in Figure \ref{fig:dft_symmetry}. In the following, we will exploit this symmetry by using the FFT to locate the valley of global optimum of $E(\omega_i;\mathcal{M}_i)$ with precision $1/T$ in a computational efficient way.

\subsection{Breaking periodicity}

\begin{figure}
    \centering
    \includegraphics[width=\linewidth]{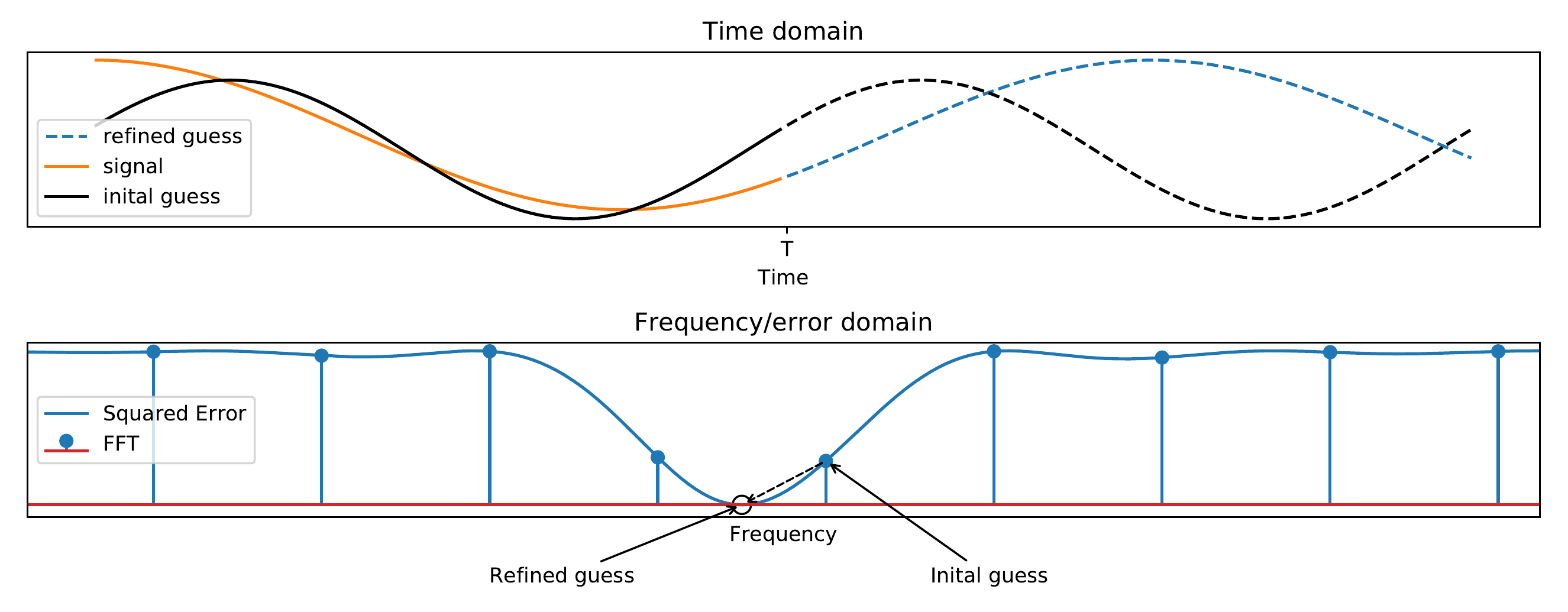}
    \vspace{-.3in}
    \caption{Graphical depiction on how the FFT and gradient descent are combined to extract global minima of the error surface. The FFT is employed to generate an initial guess of the valley in which the global optimum resides, then subsequently this initial guess is refined by gradient descent.}
    \label{fig:pfd_refine}
\end{figure}

As laid out earlier, when computing the FFT of the residual, i.e. when evaluating $E(\omega_i;\mathcal{M}_i)$ at integer multiples of $\frac{1}{T}$, implicit periodicity assumptions that are harmful for the purpose of forecasting are being made. However, we can exploit the FFT to locate the valley in which the global minimum resides and then apply gradient descent to improve this initial guess and thereby breaking periodicity assumptions. The procedure is depicted graphically in Figure \ref{fig:pfd_refine} and algorithmically in Algorithm \ref{algo:PFD}.\\

\begin{algorithm}[H]
\caption{Learning a linear oscillator from data}
\label{algo:PFD}
\begin{algorithmic}
\State Randomly initialize $\Vec{\omega}$ and $A$
\While {not converged}
\For{$i \in \{1,..,n\}$}
    \State $\omega_i \leftarrow \arg \min_{\omega_i} E(\omega_i; \mathcal{M}_i)$ (using the FFT)
    \While{not converged}
        \State $\omega_i \leftarrow \omega_i - \frac{\alpha}{T} \frac{\partial E(\omega_i; \mathcal{M}_i)}{\partial \omega_i}$ \Comment{Refine initial guess via GD}
    \EndWhile
    \State $A \leftarrow (\Omega^{T}\Omega)^{-1}\Omega^T \mathbf{X}$ \Comment{Optimize $A$ using pseudo-inverse}
\EndFor
\EndWhile
\end{algorithmic}
\end{algorithm}

The technique described in Algorithm \ref{algo:PFD} overcomes the weaknesses of the respective optimization techniques. Note that, as described earlier, while gradient descent from a random initial starting point will overemphasize temporally local losses close to $T$, the FFT will do the opposite. In order to fulfill the periodicity assumption, temporally local losses at the edges, i.e. close to $0$ and $T$, will usually be underemphasized. While gradient descent can `get stuck' in local minima and generally struggles with the fact that the sensitivity with respect to the precision of the model parameters grows linearly with time ($\sin(\omega t)' = \underline{t}\cos(\omega t)$), the FFT relies on periodicity assumptions in order to be computationally efficient. In a sense, gradient descent and the Fast Fourier Transform are local and global optimization techniques respectively and in this work, we combines global and local techniques to arrive at a global minimizer in a computationally efficient manner.

\subsection{Connection to Spectral Leakage and Bayesian Spectral Analysis}

Consider the case when $x_t$ and $y_t$ are one- and two-dimensional respectively. In that case, $R(1) = x_t$ which entails that 
\begin{align*}
E(\omega) =&  \sum_{\omega'} \hat{x}(\omega')^2 - \frac{2}{T} |\hat{x}(\omega)|^2
\end{align*}

Thus, the error of the model is closely linked to the Fourier coefficients. In a sense, this augments the traditional view of the semantics of the Fourier coefficients. In the traditional understanding, the signal is assumed to exhibit a continuous spectrum and the real and imaginary parts of Fourier coefficients denote the amplitude of the corresponding sines and cosines.

Consider the signal $x_t = cos(\omega t)$. The Fourier transform of $x_t$ is a Dirac delta function centered at $\pm \omega$. However, the discrete Fourier transform of a finite amount of equidistant samples from $x_t$ will exhibit $sinc$-shapes. This effect, usually referred to as spectral leakage~\citep{lyon2009discrete}, has been studied extensively in the signal processing community. Spectral leakage occurs when computing the FFT with finite amounts of data and can be explained in the following way: A signal of finite length can be understood as the product of an infinitely long signal and the $rect$ function stretched to $T$. Because $rect$ and $sinc$ functions are Fourier pairs and multiplication in time domain is equivalent to a convolution in frequency domain, for quasi-periodic systems, the FFT will locally exhibit the shape of the $sinc$ function. Thus, given finite amounts of data, the FFT will appear to be continuous even though the data generating process, in the limit, exhibits a discontinuous frequency spectrum. This reasoning also explains why combining the FFT and gradient descent is a promising strategy: The error surface is globally non-convex but because $sinc$ functions are suitably convex, there is local convexity. The FFT overcomes global non-convexity and gradient descent exploits local convexity.

The amount of data provided influences the width and height of the $sinc$ function. As $T$ increases, the height increases and width decreases. This entails that the more data is provided, the higher the precision with which the frequencies can be extracted from data. This view coincides and is to some degree analogous to the work by E.T. Jaynes~\citep{jaynes1987bayesian} and Bayesian Spectral Analysis~\citep{bretthorst2013bayesian}. In Bayesian Spectral Analysis, a probabilistic model of the signals spectrum is assumed and posterior inference is performed in order to obtain the most likely model parameters given the data. Interestingly, if one assumes the signal to be a simple harmonic oscillation corrupted by Gaussian noise with unknown frequency $\omega$, the DFT provides sufficient statistics of the probability that a specific frequency is present in the data. Assume $\mathbf{X}_{1:T}$ is zero-mean unit-variance and sampled at $T$ equidistant points in time and $\hat{x}(\omega)$ its Fourier transform, then it can be shown that:

\begin{align}
    p(\omega | \mathbf{X}_{1:T}, \sigma) &\propto \exp\left[ \frac{|\hat{x}(\omega)|^2}{\sigma^2}\right]\\
    p(\omega | \mathbf{X}_{1:T}) &\propto \left[ 1- \frac{2|\hat{x}(\omega)|^2}{T}\right]^{1-\frac{T}{2}}
\end{align}
with, if known, $\sigma$ denoting the standard deviation of the Gaussian noise corrupting the signal. Because the squared error is closely linked to the probability of the Gaussian distribution, this result may come as no surprise to some.

\begin{figure}[t]
    \centering
    \includegraphics[width=\linewidth]{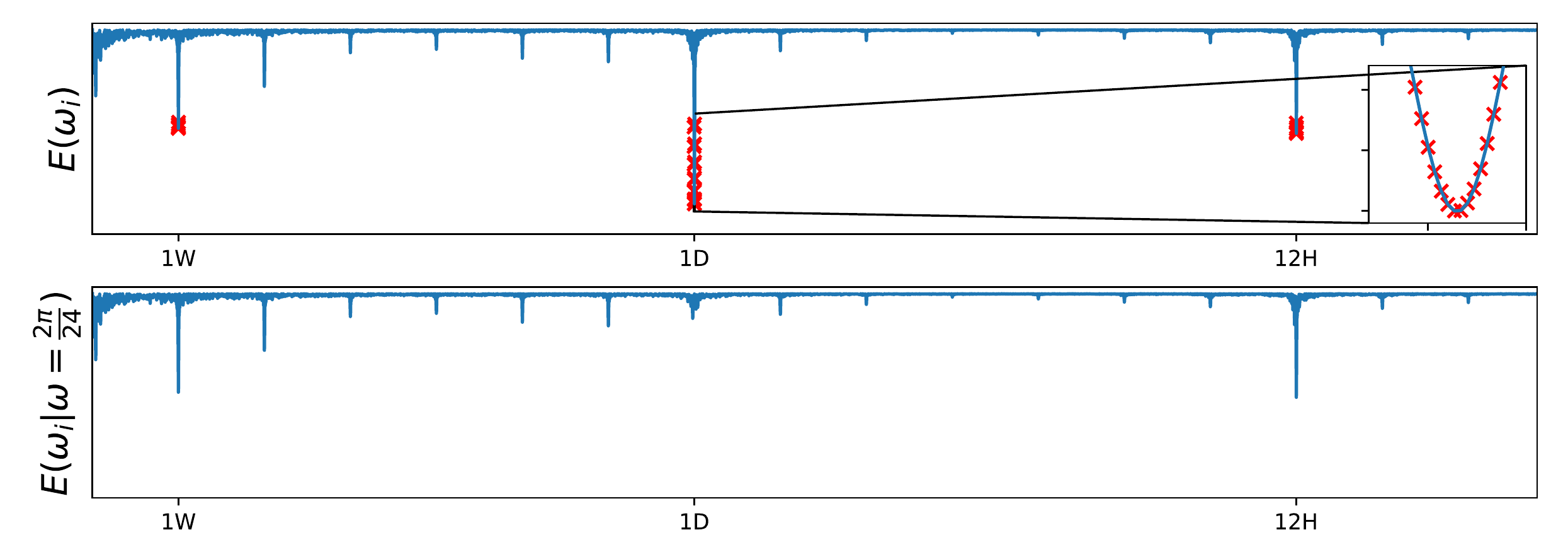}
    \vspace{-.3in}
    \caption{Top: The error surface when no previous frequencies have been selected. Red `x' mark the 50 frequencies with the lowest error. Bottom: The same error surface but when daily patterns have been explained away.}
    \label{fig:explain_away}
\end{figure}

Furthermore, note that the application of coordinate descent in the way described in Algorithm \ref{algo:PFD} has the property of `explaining away'. Because of spectral leakage, even if the true latent process exhibits a frequency spectrum that is a superposition of Dirac delta functions, the Fourier transform will exhibit $sinc$-like shapes. Thus, a na\"ive approach that computes the error surface and extracts the $k$ frequencies associated with the smallest error might select frequencies that can be explained away by a single frequency.  However, by sequentially selecting frequencies, entire $sinc$-shapes are deleted. This is shown graphically in Figure \ref{fig:explain_away}.

\subsection{Related Work}

The class of algorithms that bears the highest resemblance to the introduced algorithm is Dynamic Mode Decomposition~\citep{schmid2010dynamic} (DMD). DMD was first introduced to extract the dynamic behavior in fluid flows but has since been applied in various fields, such as epidemiology~\citep{proctor2015discovering}, acoustics~\citep{song2013global}, neuroscience~\citep{brunton2016extracting}, combustion modeling~\citep{moeck2013tomographic} and video processing~\citep{erichson2015compressed}. A strong connection between DMD and Koopman theory has been established in the past. Specifically, DMD approximates the Koopman operator when the observables are restricted to direct measurements of the state variable~\citep{rowley2009spectral,Mezic2013arfm,Tu2014jcd,Brunton2016plosone,kutz2016dynamic}.

The goal of DMD is to discover a low-dimensional linear system that captures most of the dynamic behavior of a potentially high dimensional phenomenon. Given temporal snapshots $x_t$, DMD first applies a dimensionality reduction technique and subsequently learns parameters of a linear system. Let ${\bf Z}_1$ and ${\bf Z}_2$ denote the collection of low-dimensional temporal snapshots starting at $1$ to $T-1$ and $2$ to $T$ respectively. The parameters of the linear system are usually obtained by means of pseudo-inverse, i.e. by solving ${\bf Z}_2 = {\bf A}{\bf Z}_1$. In a sense, the algorithm introduced earlier can be understood as a scalable and stable way to perform the second step of DMD, i.e. to learn parameters of the low-dimensional linear system, however by the means of FFT instead of pseudo-inverse. This is also reflected in the parameterization of the linear system. While DMD extracts a matrix $A$, our algorithm extracts a vector of frequencies $\Vec{\omega}$.

One of the main challenges for DMD-like algorithms is the fact that, under the assumption of merely measurement noise, the pseudo-inverse estimator of $\bf {A}$ is biased~\citep{hemati2017biasing,dawson2016characterizing}. Note that this bias can be explained by the fact that the pseudo-inverse estimator performs one-step-ahead predictions. This challenge has been acknowledged in the research community and approaches have been proposed to mitigate this problem, most notably with the introduction of optimized DMD \citep{chen2012variants} and de-biasing techniques~\citep{hemati2017biasing}. However, solving the optimized DMD objective proves difficult \citep{askham2018variable}. Note that the optimization objective introduced in equation (\ref{eq:mini_fourier}) is equivalent to the optimized DMD objective with the additional constraint that eigenvalues $\lambda_i$ come in conjugate pairs and are strictly imaginary which ultimately allows to obtain a globally optimal and unbiased estimator in the direction of $\omega_i$. The fact that this frequency estimator is unbiased under the assumption of merely measurement noise is easy to understand and in the following we will give the intuition why this is true but the reader is referred to pre-existing work in Bayesian Spectral Analysis~\citep{bretthorst2013bayesian} for a more thorough treatment of the issue. For a signal corrupted by measurement noise, the following holds: $x_e(t) = x(t) + \epsilon(t)$. As long as $Cov(\epsilon(t_1), \epsilon(t_2)) = 0$ and $\mathbb{E}[\epsilon(t)] = 0$, the Fourier transform of $\epsilon(t)$ is $\epsilon(t)$ which entails, because the Fourier transform is linear, that adding noise in time domain is equivalent to adding noise in frequency domain. Thus, measurement noise does not add bias but uncertainty.

Note that previously a strong connection between the Fourier Transform and DMD \citep{chen2012variants} has been shown. In \citep{chen2012variants}, the authors advocate not to subtract the mean of the data to avoid reducing DMD to the DFT to ultimately avoid the pitfalls of the DFT, i.e. making harmful and unnecessary periodicity and stationarity assumptions. In this work, enforcing stationarity is a design goal and, as pointed out earlier, periodicity assumptions are broken by the subsequent application of gradient descent in order to obtain frequencies that are not periodic in $T$. Note that the assumption of stationarity is an inductive bias and from data alone, it is usually impossible to determine whether a slow increase in the signal is caused by an upward trend or a slow frequency. Furthermore, the introduced algorithm allows to model $n$-dimensional signals composed of $m$ frequencies with $m > n$. Note that the original DMD formulation does not allow this because the dimensionality of the signal is directly tied to the number of frequencies that can potentially be extracted. However, approaches to mitigate this problem have been introduced based on mostly delay embeddings~\citep{Tu2014jcd,brunton2016extracting,brunton2017chaos,kamb2018time,pan2019structure}. Our algorithm naturally allows one to model low-dimensional signals without these additional steps.

\section{Koopman and Forecasting}

\begin{figure}[t]
    \centering
    %\vspace{-.2in}
    \includegraphics[width=\linewidth]{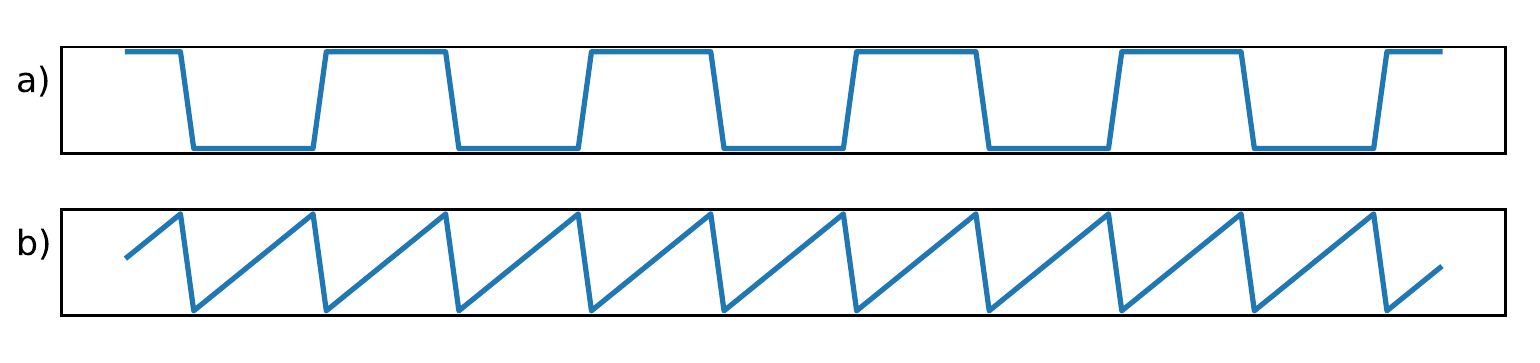}
    \vspace{-.3in}
    \caption{Two signals that both exhibit linearly decaying but infinite frequency spectra.}
    \label{fig:squarewave_koopman}
\end{figure}

Consider the two signals depicted in Figure \ref{fig:squarewave_koopman}. It is easy to show analytically that both signals exhibit linearly decaying but infinite frequency spectra. However, one could argue that both signals consist of only a single oscillatory pattern that is however not easily expressed by sines and cosines. This raises the question of whether expanding the signal in the basis of sines and cosines is the correct choice and what a better choice of oscillatory basis functions is. Note that optimal basis functions for the two signals are most likely different which implies that the choice of basis functions is dependent on the signal.\\

In the following, we will derive an algorithm based on Koopman theory that tries to answer these questions. Specifically, we will derive an algorithm that performs a frequency decomposition in a nonlinear basis. The nonlinear oscillatory basis as well as a least-squares fit is learned jointly from data. The algorithm will retain the notions of frequency and periodicity and will therefore provide some feedback to practitioners of the duration of the extracted patterns. Because the resulting algorithm is closely linked to the Fourier transform, we will show how the Fast Fourier Transform can be leveraged to obtain model parameters in a computationally efficient and scalable manner similar to approach introduced earlier, in spite of non-convexities and nonlinearities.\\
 
Koopman theory postulates that any nonlinear dynamical system can be lifted into the space of observable functions $\psi$, in which the system evolves linearly. Specifically, it can be shown that there always exists a $\mathcal{K}$ such that:
\begin{align}
    \psi(\mathbf{x}_t) = \mathbf{k}_t = \mathcal{K}\mathbf{k}_{t-1} = \mathcal{K}\psi(\mathbf{x}_{t-1}) \label{eq:koopman}
\end{align}
with $\mathcal{K}$ commonly referred to as the Koopman operator. Note that the Koopman operator and therefore $\mathbf{k}_t$ might be infinite dimensional and that $\psi$ and $\mathcal{K}$ are usually unknown.\\

Even if $\mathcal{K}$ and $\psi$ were known, forecasting $\mathbf{x}_t$ would still not be trivial because even though one could evolve $\mathbf{k_t}$ into the future, transforming knowledge of future values of $\mathbf{k}_t$ into knowledge of $\mathbf{x}_t$ is complicated. This is why in the past, $\psi$ was often assumed to be invertible. Additionally, it was assumed that $\psi$ can be approximated by a function (as opposed to a functional). In the following, we will show that these assumptions severely limit the applicability of Koopman theory. Specifically, we will show that these assumption imply that if $\mathbf{x}_t$ takes the same value for two points in time $t_1$ and $t_2$, future values $\mathbf{x}_{t_1+1}$ and $\mathbf{x}_{t_2+1}$ must be equal. Or mathematically speaking, $\mathbf{x}_{t_1} = \mathbf{x}_{t_2} \iff \mathbf{x}_{t_1 + 1} = \mathbf{x}_{t_2 + 1}$.
 \begin{align*}
    \mathbf{x}_{t_1} = \mathbf{x}_{t_2} \iff f(\mathbf{x}_{t_1}) = f(\mathbf{x}_{t_2})  \text{\quad ($f$ is function and invertible)}
\end{align*}
Evolving $f(\mathbf{x}_t)$ into the future using the Koopman operator yields:
\begin{align*}
    f(\mathbf{x}_{t_1+1}) = \mathcal{K}f(\mathbf{x}_{t_1}) =  \mathcal{K}f(\mathbf{x}_{t_2}) = f(\mathbf{x}_{t_2+1}) \iff \mathbf{x}_{t_1+1} = \mathbf{x}_{t_2+1}
\end{align*}

By making the assumption of invertibility and that $\psi$ can be approximated by a function, even very simple systems such as $x_t = \cos(t)$ cannot be lifted into a space where it evolves linearly. Let $t_1 = \frac{\pi}{2}$ and $t_2 = \frac{3\pi}{2}$, which implies that $x_{t_1} = 0 = x_{t_2}$. However, $x_{t_1+1} < 0$ and $x_{t_2+1} >0$. Note that lifting $x_t$ into a space where it evolves linearly is trivial using a functional (as opposed to function): Let $\psi(f) = \begin{bmatrix} f\\
f'
\end{bmatrix}$, then $\psi(x_t) = \begin{bmatrix} \cos(t)\\
\sin(t)
\end{bmatrix}$ which evolves linearly.\\

In order to overcome these limitations but still allow for forecasting of $x_t$, we make slightly different assumptions. In order to allow for forecasting, we assume $\psi$ to be invertible but we only assume that $\psi^{-1}$ can be approximated by a function $f$. Therefore:
\begin{align}
    \mathbf{x}_t = f(\mathbf{k}_t) &\text{ with } \mathbf{k}_t = \mathcal{K}\mathbf{k}_{t-1} \label{eq:koopman_inv2}
\end{align}

Additionally, similar to the reasoning laid out earlier, we assume $\mathcal{K}$ to have strictly imaginary eigenvalues, which in turn implies that, if $k_t$ is being projected onto the eigenvectors of $\mathcal{K}$, the following holds:

\begin{align}
    \mathbf{k}_t = \begin{bmatrix}
       \cos(\Vec{\omega} t)\\
       \sin(\Vec{\omega} t)
     \end{bmatrix}\label{eq:cos_kt} := \Omega(\Vec{\omega} t)
\end{align}

\subsection{Nonlinear Frequency Decomposition}
By making the above mentioned assumptions, we arrive at an interpretation of Koopman theory that bears resemblance to the Fourier transform. Specifically, while the Fourier transform expands a signal $x_t$ into a possibly infinite sum of sines and cosines, the transformation that results from our interpretation of Koopman theory explains the signal as a nonlinear function of sines and cosines. Thus, it explains the signal in a nonlinear oscillatory basis but because we assume the nonlinear function $f$ to be time-invariant, notions of frequency and periodicity persist, i.e., if $\Omega(\Vec{\omega} t)$ is periodic with period $P$, then $f(\Omega(\Vec{\omega} t))$ will also be periodic with period $P$.
\begin{align*}
    \mathbf{x}_t &= \mathbf{A}\Omega(\Vec{\omega} t) &\text{(Fourier)}\\
    \mathbf{x}_t &= f(\Omega(\Vec{\omega} t)) &\text{(Koopman)}
\end{align*}

Let $f$ be parameterized by $\Theta$. The minimization problem we wish to solve is therefore:
\begin{align}
    \text{minimize } E(\Vec{\omega},\Theta) &= \sum_{t=1}^T (\mathbf{x}_t - f_\Theta (\Omega(\Vec{\omega} t)))^2 & \text{(global loss)} \label{eq:koopman_loss} \\
                                            &= \sum_{t=1}^T L(\Vec{\omega},\Theta,t) \nonumber & \text{(sum of temporally local losses)}
\end{align}
In the following, we will derive computationally efficient strategies to solve this optimization problem. Again, similar to the algorithm introduced earlier, coordinate descent will be performed on (\ref{eq:koopman_loss}). We will show how a global minimizer in the direction of $\omega_i$ can be obtained in a computationally efficient manner using the FFT. Note that optimizing $E$ in the direction of $\Theta$ can be achieved efficiently using Dynamic Programming, specifically using the Backpropagation algorithm, whose implementation is trivial in programming packages that allow for automatic differentiation. However, the question of how to optimize $\Vec{\omega}$ remains.\\

\begin{figure}[]
    \centering
    \includegraphics[width=0.75\linewidth]{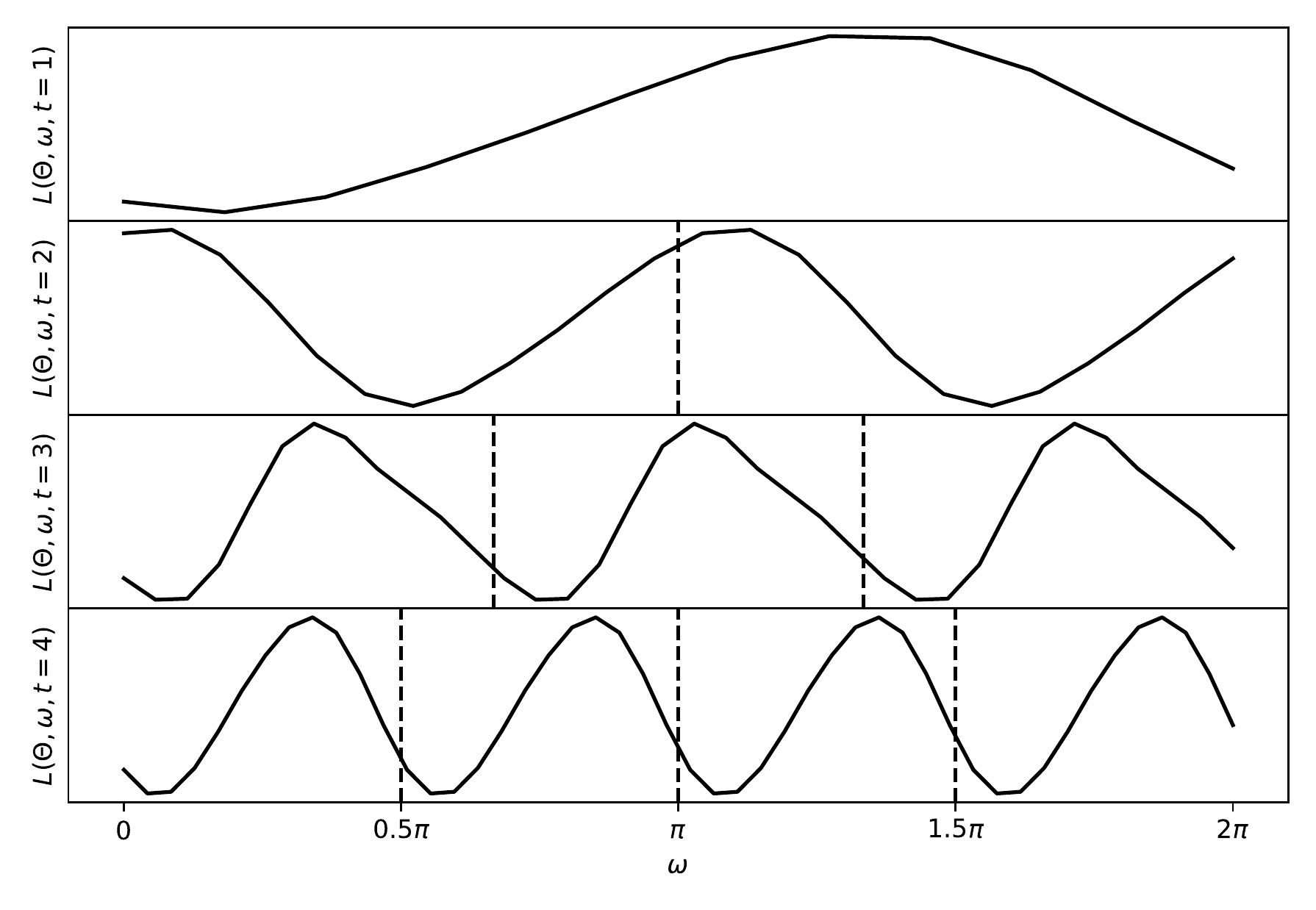}
    \caption{The temporally local loss function at time $t$ oscillates with period $\frac{2 \pi}{t}$. In order to reconstruct the global loss, it is sufficient to evaluate each local loss function within the first period.}
    \label{fig:error_periodicity}
\end{figure}

Let $\mathcal{M}_i = \{\Theta, \omega_{1}, .., \omega_{i-1}, \omega_{i+1}, .., \omega_m\}$, i.e. the set of all model parameters except $\omega_i$. Let $L(\omega_i; \mathcal{M}_i, t)$ be the loss as a function of solely $\omega_i$, i.e. while keeping all other model parameters fixed. Recall that $L(\omega_i; \mathcal{M}_i, t)$ is periodic in $\frac{2\pi}{t}$. The periodicity property follows directly from characteristics of the sine and cosine function:
$$\sin((\omega_i +\frac{2\pi}{t}) t) = \sin(\omega_i t + 2\pi) = \sin(\omega_it)$$
Thus, in a span of $2\pi$, a temporally local loss function at $t$ repeats itself $t$-times as shown graphically in Figure \ref{fig:error_periodicity}. It is therefore sufficient to compute the loss in the first period and then extrapolate this knowledge by simple repetition. Equipped with this knowledge an efficient strategy can be devised:  Each temporally local loss function is sampled at $N$ equidistant points between $0$ and $\frac{2\pi}{t}$:
\begin{align}
    \text{for }n=1,..,N \quad  s_n &= \frac{n}{N}\frac{2\pi}{t} \nonumber\\
    S_{i,t}[n] &= L(s_n; \mathcal{M}_i, t) \label{eq:loss_samples}
\end{align}

Note that this step can be carried out in parallel and is therefore surprisingly fast and that for every $i$ and $t$, $S_{i,t}[n]$ is periodic with period $N$. In order to obtain the global loss $E(\Vec{\omega},\Theta)$, in a subsequent step, the temporally local loss functions are summed up by repeating the loss at time point $t$, $t$ times and then aligning the sampling intervals. This seemingly easy step proves challenging because of the different scales of the samples drawn from the local loss functions. However, because summation requires that temporally local loss functions to be repeated and interpolated, this step can be performed efficiently in frequency domain ($\mathcal{O}(T \log T)$) by {\it zero-stuffing} and {\it zero-padding} respectively, avoiding the computational burden of a na\"ive time-domain implementation ($\mathcal{O}(T^2)$).\\

Repeating the local loss functions can be achieved efficiently by inserting $t$-many 0's in between values of the FFT of the samples drawn ({\it zero-stuffing}). Then, in order to match scales, temporally local loss functions are interpolated by appending 0's in frequency domain ({\it zero-padding}).

Therefore:
\begin{align}
    \text{for }k=0,..,K \quad & \mathbf{S}_{i,t}[k] = \sum_{n=1}^N S_{i,t}[n] e^{-j\pi\frac{nk}{K}} \label{eq:local_loss_ft}
\end{align}
For the frequency spectrum of the global loss function then holds:
\begin{align}
   \text{for } l=0,..,KT \quad \mathbf{E}_i(l) &= \sum_{t=1}^{T}\sum_{k=0}^{K}  \mathbf{S}_{i,t}[k] \delta(tk - l) \label{eq:global_loss_ft}
\end{align}

Thus, similar to the previously introduced algorithm, we derived a computationally efficient expression for the frequency spectrum of the error function which allows us to obtain the approximate location of the optimal parameter $\omega_i$. This initial guess is then refined using gradient descent. The resulting algorithm is described in Algorithm $\ref{algo:PKD}$.

\begin{algorithm}[H]
\caption{Learning a nonlinear oscillator from data}
\label{algo:PKD}
\begin{algorithmic}
\State Randomly initialize $\Vec{\omega}$ and $\Theta$
\While{not converged}
\For{$i \in \{1,..,n\}$}
    \State Compute $\mathbf{S}_{i,t}[k]$  \Comment{Based on (\ref{eq:loss_samples}) and (\ref{eq:local_loss_ft})}
    \State $E \leftarrow [0]^{TK}$
    \For{$t \in \{1,..,T\}$}
        \For{$k \in \{0,..,2N\}$}
            \State $E[tk] \leftarrow  E[tk] + S_{i,t}[k]$ \Comment Implements (\ref{eq:global_loss_ft})
        \EndFor
    \EndFor
    
    \State $\omega_i \leftarrow \arg \min_{\omega} \mathcal{F}[E](\omega)$ %E(\omega_i | \Vec{\omega}, \Theta)$ (using the FFT)
    \While{not converged}
        \State $\omega_i \leftarrow \omega_i - \frac{\alpha}{T} \frac{\partial E(\omega_i | \Vec{\omega})}{\partial \omega_i}$ \Comment{Refine initial guess of $\omega_i$}
    \EndWhile
    \For{a couple of iterations}
        \State $\Theta \leftarrow \Theta - \alpha \frac{\partial E(\Vec{\omega}, \Theta)}{\partial \Theta}$ \Comment{Gradient descent on $\Theta$}
    \EndFor
\EndFor
\EndWhile
\end{algorithmic}
\end{algorithm}

\subsection{Unknown phase problem}

\begin{figure}
    \centering
    \includegraphics[width=\linewidth]{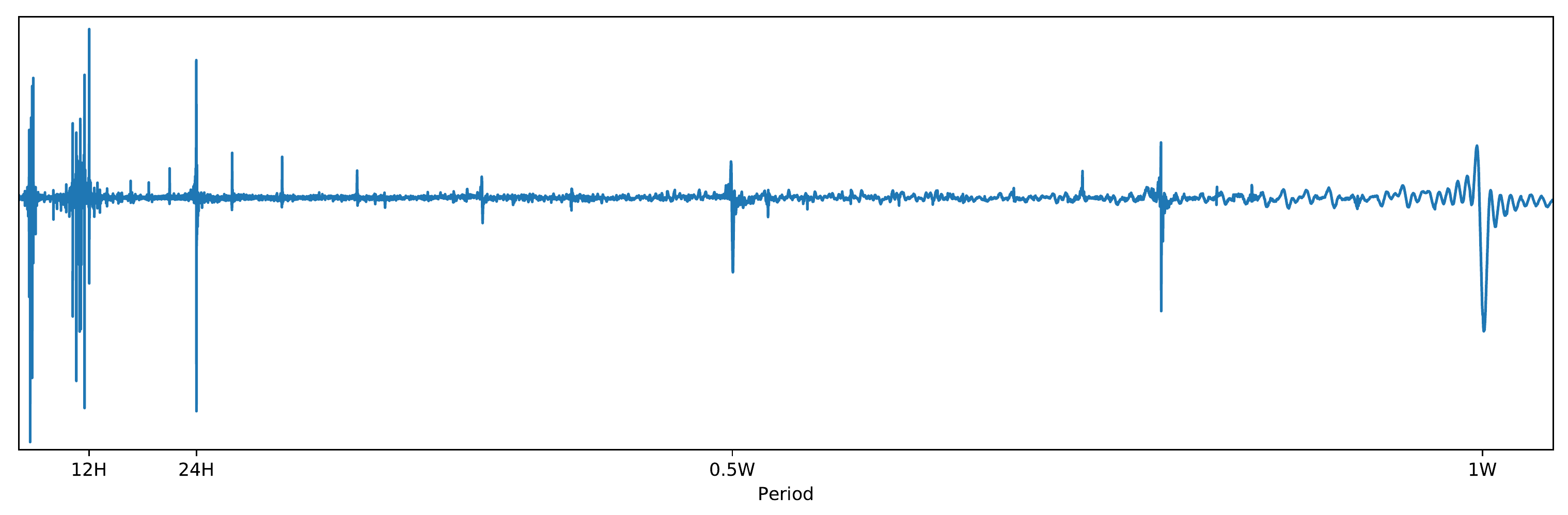}
    \caption{The error surface $E(\omega, \Theta)$ of a signal exhibiting daily, weekly and seasonal patterns. In this case $f_\Theta$ is an \emph{untrained} Neural Network.}
    \label{fig:untrained_f}
\end{figure}

The question arises to what respect the technique to compute the error surface with respect to $\omega_i$, i.e. $E(\omega_i, \mathcal{M}_i)$, relies on the non-linearity $f_\Theta$. For the Fourier-based algorithm, because the function that links the observations $\mathbf{x}_t$ and the oscillator $\mathbf{y}_t$ is linear, it can be eliminated from the optimization objective because optimal parameters can be computed analytically. Analytical solutions are unavailable for most non-linear functions $f_\Theta$ and numerically computing the optimal parameters $\Theta$ for every value that $\omega_i$ can take is computationally extremely expensive. If $f_\Theta$ is approximated by a Neural Network, it would require training a Neural Network to convergence hundreds of thousands of times. One might ask: Is it possible to extract meaningful frequencies $\omega$ even with a suboptimal $f_\Theta$? Surprisingly, even with an untrained Neural Network, i.e. with random weights, meaningful frequencies can be extracted. Figure \ref{fig:untrained_f} shows the error surface with respect to a signal that exhibits daily, weekly and seasonal patterns. It is easy to see that even with an untrained Neural Network, the correct frequencies can be extracted. However, there is a price to pay for not knowing the optimal $\Theta$ for every $\omega_i$, namely that the correct frequency merely cause the error function to deviate from its plateau. Thus, even an upward spike in the error function can indicate a correct frequency. We call this issue the \emph{unknown phase problem}. Training the Neural Network for some iterations or rerunning the experiments from different initial conditions often alleviates this problem.

\subsection{Related Work}

The fact that any nonlinear dynamical system can be lifted by the means of observables into a space where its time evolution can be described by linear methods was first introduced by Koopman for Hamiltonian systems~\citep{koopman1931hamiltonian} and later generalized to continuous-spectrum systems \citep{koopman1932dynamical}. Even then it was of considerable importance as a building block for advances in ergodic theory~\citep{birkhoff1931proof,neumann1932proof,birkhoff1932recent,neumann1932physical,moore2015ergodic}. Interest in Koopman theory was renewed by work by \cite{mezic2005spectral,budivsic2012applied,Mezic2013arfm}. More recent theoretical advances in Koopman theory e.g. include generalization of the Hartman-Grobman theorem to the entire basin of attraction
of equilibrium points or periodic orbits \citep{lan2013linearization}.

Recent applied research has focused on algorithms to estimate the Koopman operator from data. Most of these approaches rely on auto-encoder structures~\citep{otto2019linearly,lusch2018deep,yeung2019learning,wehmeyer2018time,takeishi2017learning}. The optimization objective is usually constituted of terms that encourage linearity in the latent space as well as reconstruction performance and is usually solved by means of gradient descent. These approaches have been extended in various ways. For example, Bayesian Neural Networks as encoders were utilized to extend Koopman theory to the probabilistic setting~\citep{pan2019physics} whereas \cite{champion2019data} relaxed the linearity requirement and allowed for sparse dynamics in the latent space. These approaches usually do not consider linearity in Koopman-space a constraint but rather the optimization objective. This usually results in latent spaces that are approximately and only locally linear. On top of that, unstable eigenvalues often cause difficulties. In~\cite{erichson2019physics}, stable eigenvalues are encouraged through additional regularization by means of Lyapunov stability. In this work, on the other hand, global linearity and stability are constraints of the optimization objective, i.e. all solutions are globally linear and stable by construction without the need for any additional regularization. Furthermore, in this work, we show that the global optimization objective is highly non-convex limiting the practicality of gradient descent and show how solutions can be obtained in frequency domain.

\section{Algorithmic properties}

In this section, basic properties of the algorithms are derived.

\subsection{Universal approximation capabilities}

\subsubsection{Finite data}
We will show that even the linear Fourier-based algorithm has universal approximation capabilities for finite data sets. More formally, we show by construction that there always exists finite model parameters that can explain any finite data set. Note that this property is not very useful in practice because the solution does not allow to generalize as it implicitly assumes that the signal is periodic in $T$. It is included here because it is analogous to Cover's theorem. Similarly to Cover's theorem, colloquially speaking, we will show that even the linear Fourier-based algorithm can always overfit on the training set. Analogous to Cover's theorem, the proof requires projecting the data into an $N$-dimensional space. In summary, Cover's theorem and the Fourier-based algorithm both have trivial solutions in $N$ dimensions that however do not generalize.

\begin{lemma}[Finite data: Universal approximation capabilities]
Assume $T$ equidistant samples $x_1, ..., x_T$. There always exist finite-dimensional $\Vec{\omega}$ and $A$ such that, $A\Omega(\omega t) = x_t$. Proof by construction: Choose $M=T/2$ and for $i=1,..,M$ let $\omega_i = \frac{2\pi i}{T}$. Let $O \in \mathbb{R}^{T \times T}$ the matrix whose $i$th and $(i+T/2)$th rows constitute $\cos(\omega_i t)$ and $\sin(\omega_i t)$ evaluated at $t = 1,..,T$, respectively. What is left to show is that $O$ forms a basis of $\mathbb{R}^T$ which follows directly from 
\begin{align*}
i \neq j: \sum_{t=1}^T \sin(\frac{2\pi it}{T})\sin(\frac{2\pi jt}{T}) = 0\\
i \neq j: \sum_{t=1}^T \cos(\frac{2\pi it}{T})\cos(\frac{2\pi jt}{T}) = 0\\
i \neq j: \sum_{t=1}^T \sin(\frac{2\pi it}{T})\cos(\frac{2\pi jt}{T}) = 0
\end{align*}

\end{lemma}

\subsubsection{Infinite data}
However, when the signal is assumed to be infinite, we will show that the Koopman-based algorithm is more expressive compared to its Fourier counterpart. Specifically, we will show that there are signals for which the Fourier algorithm would require infinite model size, while the Koopman algorithm allows for a finite parameterization. The signal for which this is shown is a simple square wave.

\begin{lemma}[Infinite data]
Let $\%$ denote the modulo operator and consider the time-discrete system that evolves according to the following equations: 
\begin{align*}
 x_t=
\begin{cases}
1, \text{ for } t\%4 =0\\
1, \text{ for } t\%4 =1\\
0, \text{otherwise}
\end{cases}
\end{align*}

It is well known that the discrete-time Fourier transform of $x_t$ is linearly decaying but infinite. The system can however be modeled with a simple nonlinear function $f$ and a single frequency:

\begin{align*}
 f(x)=
\begin{cases}
1, \text{ for } x > 0\\
0, \text{otherwise}
\end{cases}\\
x_t = f(\cos(\frac{\pi}{2}t))
\end{align*}

\end{lemma}

\subsection{Error bounds}

In the following, practical error bounds will be derived as a function of noise and amount of data provided.

\begin{lemma}[Linear-in-time error bounds]
\label{lemma:lit}
Let $\overline{x}_t(\Vec{\omega}) = f(\Omega(\Vec{\omega}t))$. Assume $f$ is scalar-valued and Lipschitz continuous with constant $\lambda$. Note that $\Omega(\Vec{\omega}t)$ is Lipschitz continuous with constant $t$, because $\frac{\partial \Omega(\Vec{\omega}t)}{\partial \Vec{\omega}} = t \begin{bmatrix}
       -\sin(\Vec{\omega} t)\\
       \cos(\Vec{\omega} t)
     \end{bmatrix}$ and $\sin$ as well as $\cos$ are bounded by $[-1,1]$. Because any composition $f \circ g$ is Lipschitz continuous with constant $L=AB$ for all $f$ and $g$ that are Lipschitz continuous with constants $A$ and $B$ respectively, $\overline{x}_t$ is Lipschitz continuous with constant $\lambda t$.\\

Because $\overline{x}$ is Lipschitz continuous with constant $\lambda t$, we know that:
\begin{align*}
    |\overline{x}_t(\Vec{\omega}) - \overline{x}_t(\Vec{\omega}^*)| \leq \lambda t \sum_i|\omega_i - \omega_i^*|
\end{align*}

From Bayesian Spectral Analysis, we know the standard deviation of the uncertainty with which the frequencies can be resolved. Let $A$ and $\sigma^2$ denote amplitude and noise respectively. As Jaynes derived:

\begin{align*}
    \omega = \omega^* \pm \frac{\sigma^2}{A} \sqrt{\frac{48}{T^3}}
\end{align*}

Assume $\omega_i$ was resolved within one standard deviation of $ \omega_i^*$, then:
\begin{align*}
\omega_i^* - \frac{\sigma^2}{A_i} \sqrt{\frac{48}{T^3}} \leq \omega_i \leq \omega_i^* + \frac{\sigma^2}{A_i} \sqrt{\frac{48}{T^3}}\\
    |\omega_i - \omega_i^*| \leq \frac{\sigma^2}{A_i} \sqrt{\frac{48}{T^3}}
\end{align*}
    Combining results yields:
\begin{align*}
    |\hat{x}_t(\Vec{\omega}) - \hat{x}_t(\Vec{\omega}^*)| \leq \lambda t \sum_i|\omega_i - \omega_i^*| \leq \frac{\lambda t}{\sqrt{T^3}} \sum_i \frac{\sqrt{48} \sigma^2}{A_i} \\
    |\hat{x}_t(\Vec{\omega}) - \hat{x}_t(\Vec{\omega}^*)| \in \mathcal{O}\left(\frac{t}{\sqrt{T^3}} \sum_i \frac{\sigma}{A_i}\right)
\end{align*}

\end{lemma}

Thus, the algorithm scales super-linearly with the amount of data provided and degrades linearly with noise variance and prediction horizon $t$.

\section{Experiments}

The code for the Fourier- and Koopman forecasting algorithms is available at \url{https://github.com/helange23/from_fourier_to_koopman}.

\subsection{Synthetic experiments}

In the following, the efficacy the algorithms is probed on synthetic tasks and compared to the Long-Short Term Memory (LSTM)~\citep{hochreiter1997long}. LSTM is a trainable nonlinear filter and one of the de facto standard tools to model time series. LSTMs were initially developed to mitigate the problem of exploding or vanishing gradients by choosing a specific neural architecture that ensures constant error flow. Note that exploding gradients is also a problem we faced but instead mitigated by employing the FFT.

\subsubsection{Forecasting nonlinear oscillators under measurement noise}

In order to probe the performance of the resulting algorithms in long-term forecasting of nonlinear oscillators under measurement noise, a synthetic time series is created. Specifically, we choose a strongly nonlinear signal resembling data at an hourly rate with daily patterns that is corrupted by Gaussian noise, i.e.:
\begin{align*}
    x_t = \sin \left(\frac{2\pi}{24} t \right)^{17} + \epsilon_t
\end{align*}

In a first experiment, we vary the amount of training data provided to the models and keep the noise variance fixed at $0.2$. Each model is provided with 2000, 4000, 6000 and 8000 samples of $x_t$ and the performance at predicting the next $10^4$ snapshots is evaluated. For the LSTMs, a 5-layer 20-unit network comprising 15301 parameters was chosen. For long-term predictions, previously predicted snapshots are recursively fed back into the network. The LSTM is compared to the Koopman algorithm with a 3-layer decoder with roughly the same number of parameters (16384) and a single frequency. The Fourier algorithm was instantiated with 8 frequencies and therefore 16 parameters.

\begin{figure}[t]
    \centering
    \includegraphics[width=\linewidth]{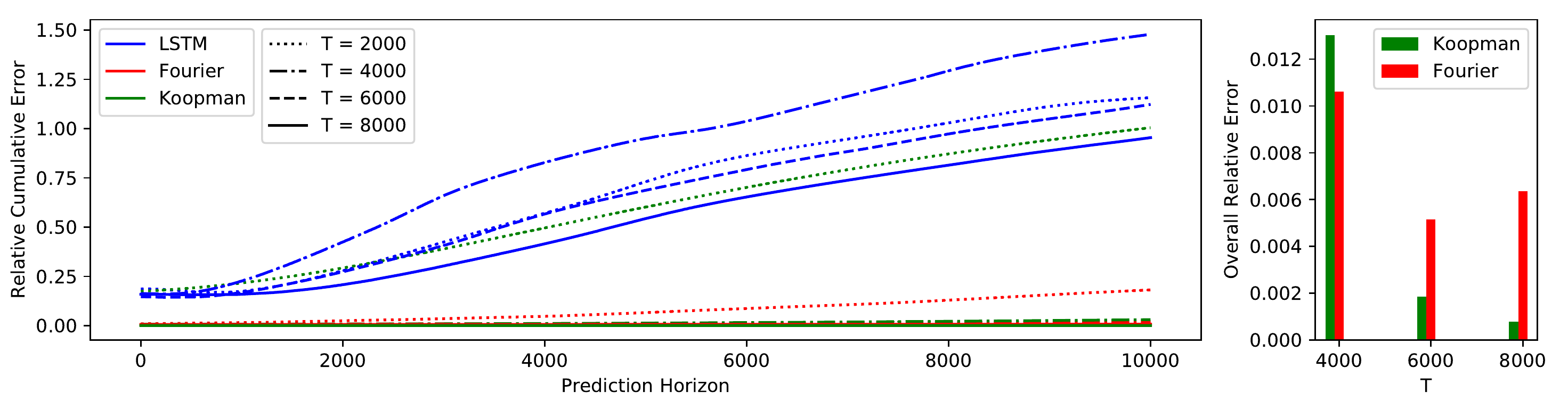}
    \vspace{-.2in}
    \caption{Left panel shows the Relative Cumulative Error as a function of the amount of training data provided. The right panel displays the Relative Cumulative Error when the prediction horizon equals $10^4$ but, for better visibility, leaves out the results for the LSTM and T=2000.}
    \label{fig:res_T_nonlinear}
\end{figure}

The left panel of Figure \ref{fig:res_T_nonlinear} shows the Relative Cumulative Error as a function of the prediction horizon and the amount of provided training data. We define the Relative Cumulative Error as:
\begin{align}
    RCE(h) = \frac{\sum_{t=1}^{T+h}(x_t - \hat{x}_t)^2}{\sum_{t=1}^{T+h} x_t^2}
\end{align}

\begin{figure}
    \centering
    \includegraphics[width=\linewidth]{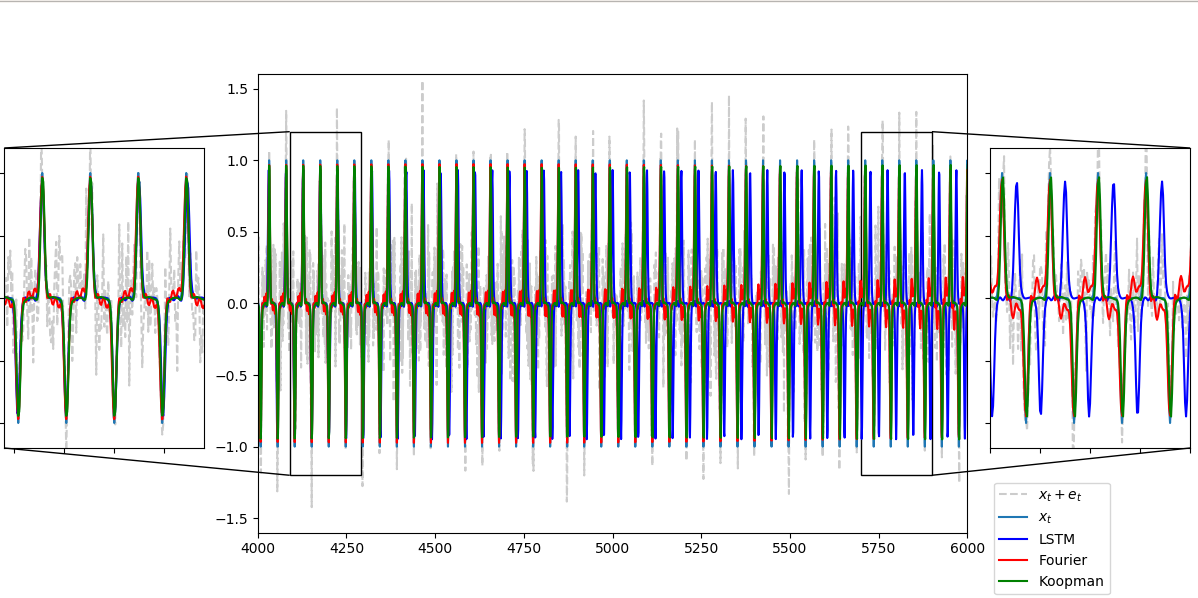}
    \vspace{-.2in}
    \caption{Long-term forecasting results for $T=4000$. Note that LSTMs correctly extract the shape of the nonlinearity but fail to extract the correct frequency. Furthermore note that for the Fourier algorithm small Fourier modes seem to be drowned by noise.}
    \label{fig:T_compare}
\end{figure}

For the sake of readability, the right panel shows the performance of only the Fourier-based and Koopman-based algorithms on a subset of the benchmarks when the prediction horizon equals $10^4$. Note that this experiment confirms the theoretical findings provided earlier, namely that the algorithms scale super-linearly with the amount of data. Furthermore, note that the Koopman algorithm outperforms the Fourier algorithm. Figure \ref{fig:T_compare} gives the intuition why this is true.\\

Figure \ref{fig:T_compare} shows the time-domain long-term forecasts of the respective algorithms. Note that the LSTM seems to have correctly extracted the nonlinear shape of the waveform but failed at extracting the correct frequency. The impression that LSTMs seem to be biased frequency estimators will be reinforced in later experiments. Furthermore note that the Koopman algorithm significantly outperforms the Fourier algorithm. While the Fourier algorithm correctly extracted large Fourier modes from the data, it struggles extracting smaller ones as they appear to be wiggling slowly out of phase. This behavior can be explained by the theoretical findings earlier: smaller Fourier modes have a smaller signal-to-noise-ratio compared to bigger ones and are therefore harder to detect correctly.\\

\begin{figure}
    \centering
    \includegraphics[width=\linewidth]{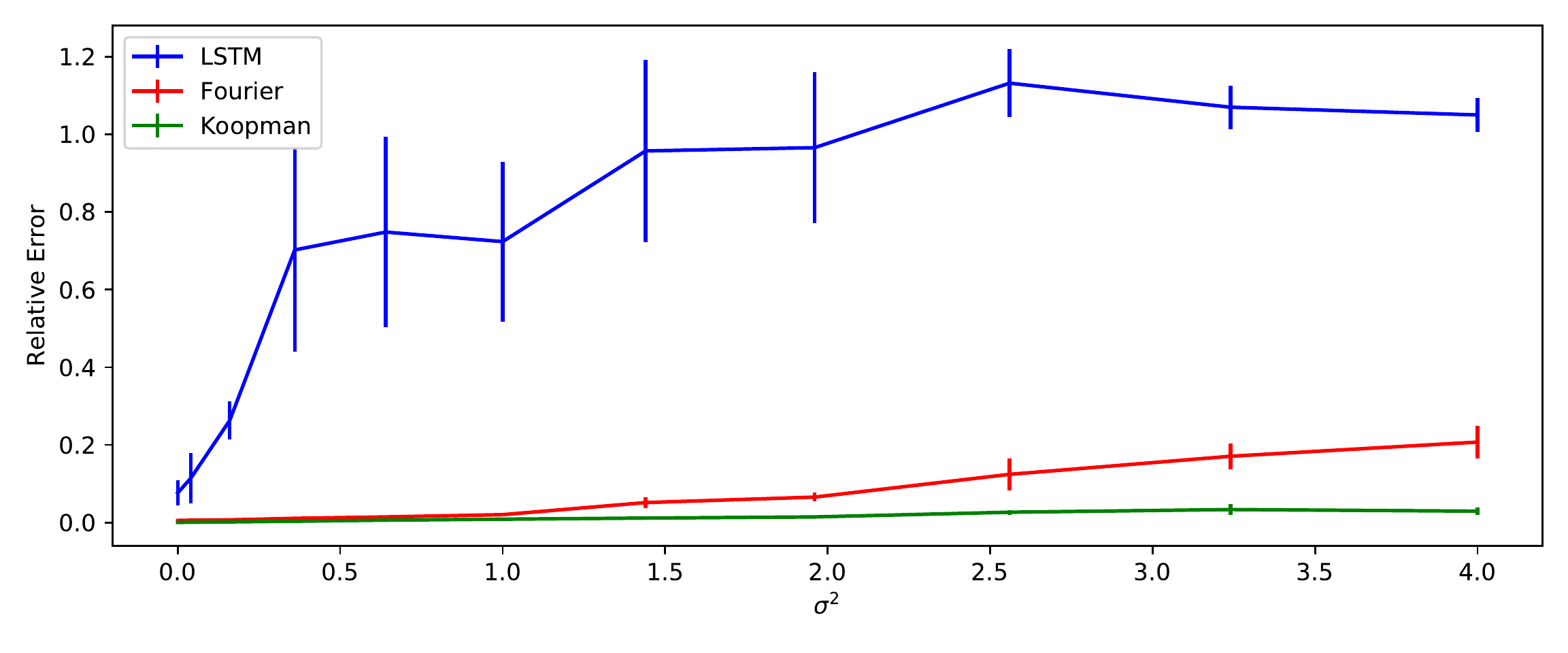}
    \vspace{-.4in}
    \caption{Robustness of the respective algorithms to noise. The amount of training data is kept fixed at $T=4000$ while noise variance is varied. The Koopman-based algorithm significantly outperforms its Fourier counterpart as well as LSTMs.}
    \vspace{-.2in}
    \label{fig:noise_nonlinear}
\end{figure}

In a subsequent experiment, we probe the robustness of the respective algorithms to noise by keeping the amount of training data fixed ($T=4000$) and varying the variance of the Gaussian noise corrupting the signal. Figure \ref{fig:noise_nonlinear} shows the results. First, note that the theoretical results are again confirmed, i.e. the performance of the algorithms seems to degrade linearly with noise variance. Furthermore note that, again, the Koopman-based algorithm outperforms its Fourier counterpart. This can, again, be explained by the fact that the Fourier-based algorithm requires more Fourier modes to model the signal appropriately. However, these additional Fourier modes often have small amplitudes and therefore small signal-to-noise ratios and get therefore drowned out by noise quickly. The Koopman-based algorithm avoids this pitfall as it requires fewer frequencies because it can explain away harmonics using the non-linearity.

\subsubsection{Traveling waves}

For many data-driven modeling approaches to high dimensional systems such as e.g. fluid flows, the dimensionality of the measurements are often first reduced by applying dimensionality reduction techniques. The Koopman-based algorithm allows for jointly learning the inverse of the transformation that reduces the dimensionality and therefore extends the applicability of the techniques to high dimensional systems with which traditional dimensionality reduction techniques such as Principal Component Analysis (PCA) struggle. A standard example for systems with which PCA struggles are traveling waves. Because PCA exploits redundancies in the co-variance structure of the measurements and measurements within one cycle of a traveling wave are approximately uncorrelated, achieving good reconstruction with PCA requires many modes which in turn reduces robustness to noise as highlighted in previous experiments. In the following, we will show how the Koopman-based algorithm can overcome these difficulties. Specifically, by showing how a 2D traveling wave can be modeled with a transpose-convolutional decoder.

\begin{figure}
    \vspace{-.2in}
    \centering
    \includegraphics[width=\linewidth]{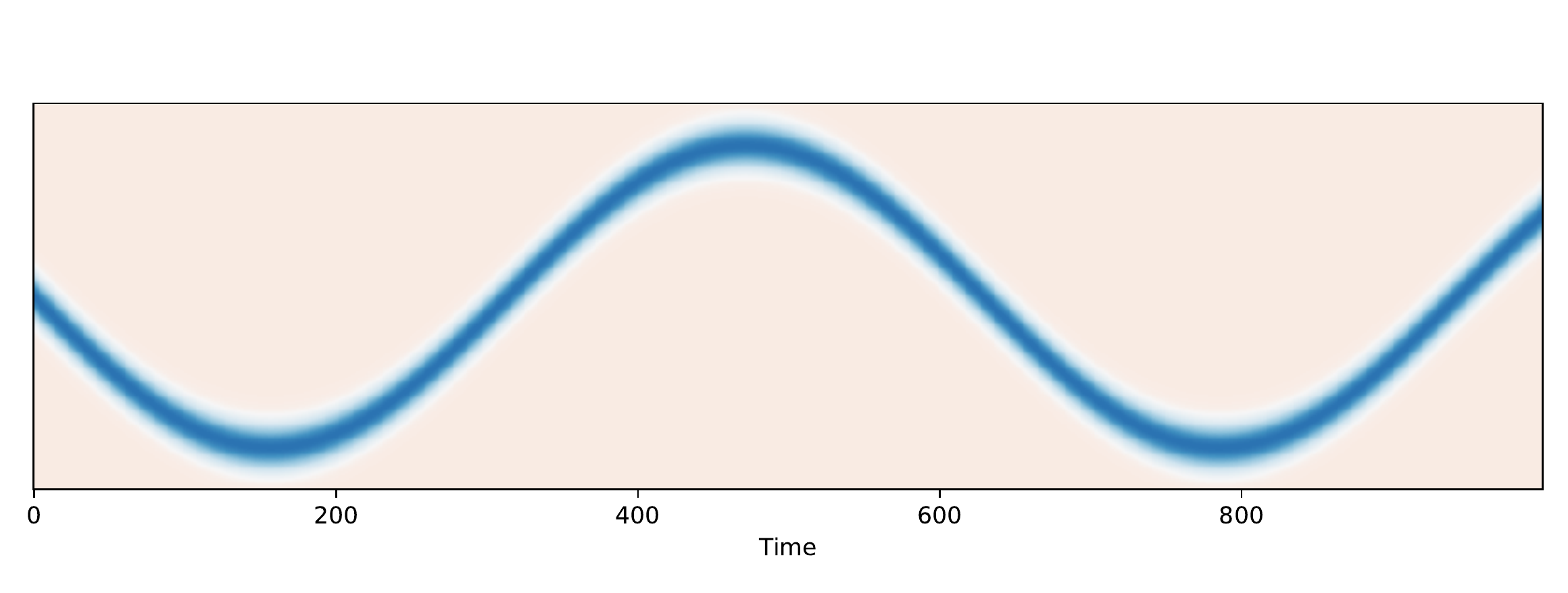}
    \vspace{-.4in}
    \caption{A traveling wave for which reducing the dimension with PCA proves difficult.}
    \label{fig:travel_wave}
    %\vspace{-.2in}
\end{figure}

\begin{figure}[t]
    \centering
    \includegraphics[width=\linewidth]{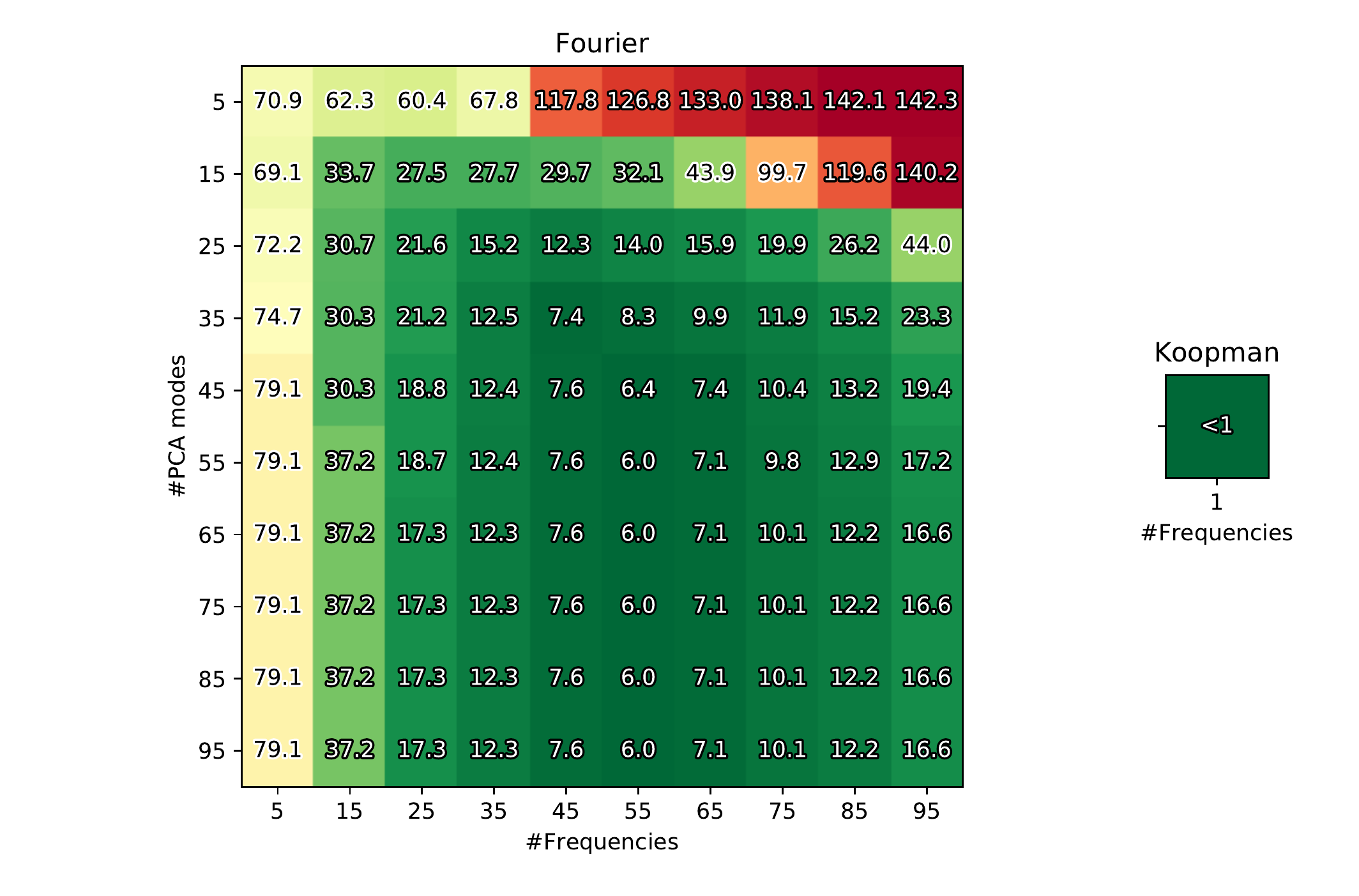}
    \vspace{-.4in}
    \caption{The results of a parameter sweep of the Fourier-based algorithm in conjunction with PCA on modeling a traveling wave in comparison to the Koopman-based algorithm with a deconvolutional decoder.}
    \label{fig:travel_wave_results}
        \vspace{-.2in}
\end{figure}

Figure \ref{fig:travel_wave} shows the first 1000 snapshots of a traveling wave with spatial dimensionality of 256. Let each spatial dimension be indexed by $1 \leq u \leq 256$. For the spatio-temporal evolution of the wave holds: $$x(u,t) = \mathcal{N}(u| \mu = (\sin(0.01 t) + 1)*100 + 28, \sigma^2 = 10)$$ The performance of the Fourier-based algorithm in conjunction with PCA is compared to an instance of the Koopman-based algorithm that employs a transpose-convolutional Neural Network as a decoder. For the Fourier-based algorithm in conjunction with PCA, a parameter sweep over the combination of retained PCA modes and the number of frequencies is performed. In total, 100 combinations are tested and the results are shown in Figure \ref{fig:travel_wave_results}. The best performing model incurs an error of approximately 6\% with 55 retained PCA modes and frequencies, respectively. On the other hand, the Koopman-based algorithm in conjunction with a transpose-convolutional decoder achieves an error smaller than 1\% with just a single frequency.

\subsection{Natural data experiments}

\subsubsection{Energy demand: RE-Europe data set}
In the following, we will show that physical systems can be modeled effectively as nonlinear oscillators. The example at hand is taken from the realm of power systems. Specifically, the goal of the experiment is to predict the total energy consumption at the distribution level of a node in the German electrical grid. Predicting future consumption is paramount for numerous applications such as power quality, grid stability and demand response~\citep{chan2012load}. For this, a one-dimensional time series of past demand is extracted from the RE-Europe data set~\citep{jensen2017re}. Energy demand usually exhibits multiple scales, i.e. because energy consumption is usually higher during weekdays compared to weekends, energy consumption often exhibits weekly alongside daily and seasonal patterns. On top of that, because energy consumption is dependent on hard-to-predict outside factors such as weather, predicting energy demand into the future requires high robustness to structured noise. The data set at hand contains 3 years of data at an hourly rate, therefore 26.280 data points. The first 19.000 data points were used as training, the next 1000 for testing, and the last 6.280 for evaluation. The evaluation set was in turn divided into four equal in size consecutive subsets. The performance of the Fourier- as well as Koopman-based algorithm is compared to various temporal models such LSTM, GRU, Box-Jenkins models, Clockwork-RNNs, Echo State Networks and Fourier Neural Networks.\\

\begin{table}[]
\centering
\begin{tabular}{|l||c|c|c|c||c|c|c|}
\hline
Algorithm                       & \multicolumn{4}{c ||}{Forecast Horizon} & \multicolumn{3}{c |}{Patterns} \\
\hline
\hline
                                & 25\%    & 50\%    & 75\%   & 100\%   & D        & W       & Y       \\
\hline
\hline
Koopman Forecast                    &  $\mathbf{0.19}$ &  $\mathbf{0.21}$ & $\mathbf{0.19}$ & $\mathbf{0.19}$ &  \checkmark  & \checkmark & \checkmark \\
\hline
Fourier Forecast                         &  0.31  &  0.39  & 0.33 &  0.3  & \checkmark & \checkmark  &  \checkmark \\
\hline
LSTM                            &  $0.37$  & 0.4 & 0.42  &  0.45  & \checkmark  & $\times$  &  $\times$  \\
\hline
GRU                           &  0.53  &  0.55  & 0.52  & 0.5 & \checkmark &  $\times$ &  $\times$ \\
\hline
Echo State Network         &  0.67   & 0.73 &  0.76  & 0.73    & \checkmark & $\times$ & $\times$ \\
\hline
AR(1,12,24,168,4380,8760)   & 0.75  &  0.95  & 1.07  &  1.13  & \checkmark & \checkmark & \checkmark  \\
\hline
CW-RNN (data clocks)                      &  1.1  &  1.14  &  1.14  & 1.15 & (\checkmark)  & $\times$ & $\times$ \\
\hline
CW-RNN                          &  1.05  &  1.08  & 1.08  & 1.09    & (\checkmark)  & $\times$  & $\times$ \\
\hline
AutoARIMA               & 0.83 & 1.11  & 1.18    & 1.26 & $\times$ & $\times$ & $\times$ \\
\hline
Fourier Neural Networks         &  1.1  &  1.15  & 1.21   & 1.21 &  \checkmark &  $\times$ &  $\times$ \\
\hline
\end{tabular}
\label{tab:results_energy11}
\caption{Performance in long-term forecasting of distribution level energy consumption as measured by the Relative Cumulative Error for various algorithms. Note that long-term predictions are obtained by recursively feeding predictions back into algorithms where applicable. Furthermore, column Patterns indicates whether algorithms at hand have successfully extracted \textbf{D}aily, \textbf{W}eekly or \textbf{Y}early patterns.}
\end{table}

%\begin{figure}
%    \centering
%    \includegraphics[width=\linewidth]{plots/energy_lstm.pdf}
%    \caption{Reconstruction and generalization performance of various configurations of LSTMs. Long-term predictions are obtained by recursively feeding one-step-ahead prediction backk into the network. Best performing model correctly explains 55.2\% of the variance.}
%    \label{fig:res_energy_lstm}
%\end{figure}

For LSTMs and GRUs, the standard pytorch implementation is utilized and a parameter sweep over the number of layers as well as the number of hidden units per layer is performed totalling 100 different configurations. For the Box-Jenkins models the Hyndman-Khandakar~\citep{hyndman2007automatic} AutoARIMA algorithm implemented in the \emph{R} forecasting package is employed which uses a combination of unit root tests, minimization of the Akaike Information Criterion~\citep{sakamoto1986akaike} and Maximum Likelihood Estimation to obtain an ARIMA model. We compare to the predicted mean. Note that the hyperparameter space for Box-Jenkins models is very large and computationally too expensive to sweep. But in addition to the AutoARIMA approach, handcrafted ARIMA models are created. Because of the long-range dependencies in the data, the best practices of detrending and differencing followed by an investigation of (partial) auto-correlation functions failed. That is why numerous hand-crafted models imbued with expert knowledge of the underlying data are probed. The best performing hand-crafted model is a simple auto-regressive process with lags corresponding to periods and harmonics found in the underlying data. Additionally, the performance of Clockwork-RNNs~\citep{koutnik2014clockwork} (CW-RNN) is evaluated. CW-RNNs have an internal representation of periodicity and frequency because subsets of the internal state are updated based on fixed (not learned) periods of internal clocks. We compare to a CW-RNN whose internal clock is instantiated according to the power law as described by the authors but also compare to a CW-RNN whose internal clock ticks according to the periods found in the data. For both models, a single layer CW-RNN was utilized and a parameter sweep over the number of clocks and states per clock was performed. Adjusting the internal clock to the underlying data did not improve the model. Furthermore, we compare to Fourier Neural Networks~\citep{silvescu1999fourier,gashler2016modeling} and Echo State Networks~\citep{jaeger2007echo}. Fourier NNs are by design not temporal models but merely employ $\cos$-activation functions. Temporal dependencies were attempted to be modeled by learning a function that produces the value of the next time step as a function of the previous 10. For Echo State Networks, the 'easyesn' package was employed. Table \ref{tab:results_energy11} shows the results.

%\begin{figure}
%    \centering
%    \includegraphics[width=\linewidth]{plots/energy_fourier.pdf}
%    \caption{Reconstruction and generalization performance of the Fourier-based algorithm as a function of extracted frequencies. Best performing model correctly explains 77.4\% of the variance.}
%    \label{fig:res_energy_fourier}
%\end{figure}

%Figures \ref{fig:res_energy_fourier} and \ref{fig:res_energy_koopman} show the performance of the Fourier- and Koopman-based algorithms respectively. The Koopman algorithm was instantiated with a fully-connected three-layer perceptron as the decoder. The number of hidden units was varied in the experiments, while input dimensionality is predetermined by the number of frequencies. The Fourier and Koopman methods significantly outperform LSTMs in long-term predictions, whereas the Koopman-based algorithm slighlty outperforms its Fourier counterpart.

%\begin{figure}
%    \centering
%        \includegraphics[width=\linewidth]{plots/energy_koopman.pdf}
%    \caption{Performance of the Koopman algorithm as a function of the number of hidden units in the decoder and number of extracted frequencies. Best performing model correctly explains 83.9\% of the variance.}
%    \label{fig:res_energy_koopman}
%\end{figure}

\begin{figure}[t]
    \centering
    \includegraphics[width=\linewidth]{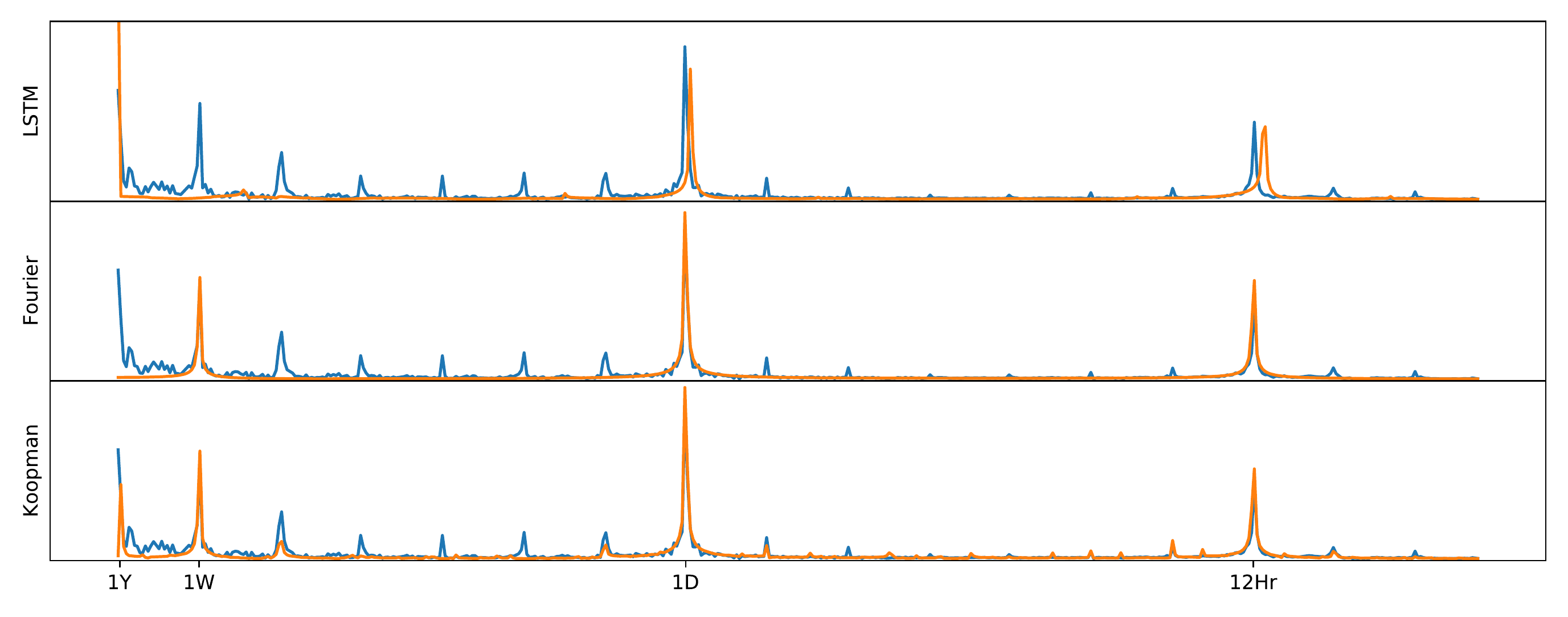}
    \vspace{-.3in}
    \caption{Comparison of the best performing model of each contender in frequency domain. The x-axis shows frequencies with periods of one year, one week, one day and 12 hours. The LSTM fails to extract slow frequencies and seems to be a biased frequency estimator.}
    \label{fig:res_energy_fft_compare}
    \vspace{-.2in}
\end{figure}

Figure \ref{fig:res_energy_fft_compare} exemplifies the pitfalls of time-stepping methods for long-term time series prediction. It shows the frequency content of the prediction of the Koopman- and Fourier-based algorithm alongside the prediction of the best performing LSTM. In all experiments, time stepping methods like LSTMs, GRUs or ESNs have two important limitations. First, because their objective is one-step-ahead prediction and slow frequencies incur little error in the short-term, they fail to extract slow frequencies from data. Second, the resulting predictions are biased in terms of frequency. When integrating the LSTM (recursively feeding predictions back into the network), the resulting time series does oscillate but not with the correct frequency. This behavior is detrimental, especially when the Mean Squared Error is employed to evaluate the performance. Predicting a constant often incurs a smaller error then predicting an out-of-phase oscillation.

\subsubsection{Fluid flows and Video Frame Prediction}

There seems to be a dichotomy in fluid flows. On one hand, there are noise amplifying flows and on the other hand there are oscillator flows. Because of the inductive bias of the algorithms introduced in this work, we will focus on evaluating the performance of the proposed algorithms in forecasting the behavior of oscillator flows. The performance is compared to Dynamic Mode Decomposition (DMD) and a variant of DMD designed to be more noise robust, namely the forward-backward DMD algorithm~\citep{hemati2017biasing}. Note that in the following experiments, the dimensionality of the high dimensional flows was reduced by Principal Component Analysis and the temporal forecasting was performed using the Fourier-based algorithm. For each experiment, the first 75\% of the data was used for training whereas the remaining 25\% are used for testing (temporal split).\\

\begin{table}[h!]
\begin{tabular}{|cc|c|c|c|}
\hline
                                        & & \textbf{Prediction} & \textbf{Ground Truth} \\
\hline
\rotatebox[origin=l]{90}{\textbf{Kolmogorov 2D}}  & \rotatebox[origin=l]{90}{Rey. no.: $\approx 10$\ } &   \includegraphics[width=0.4\linewidth]{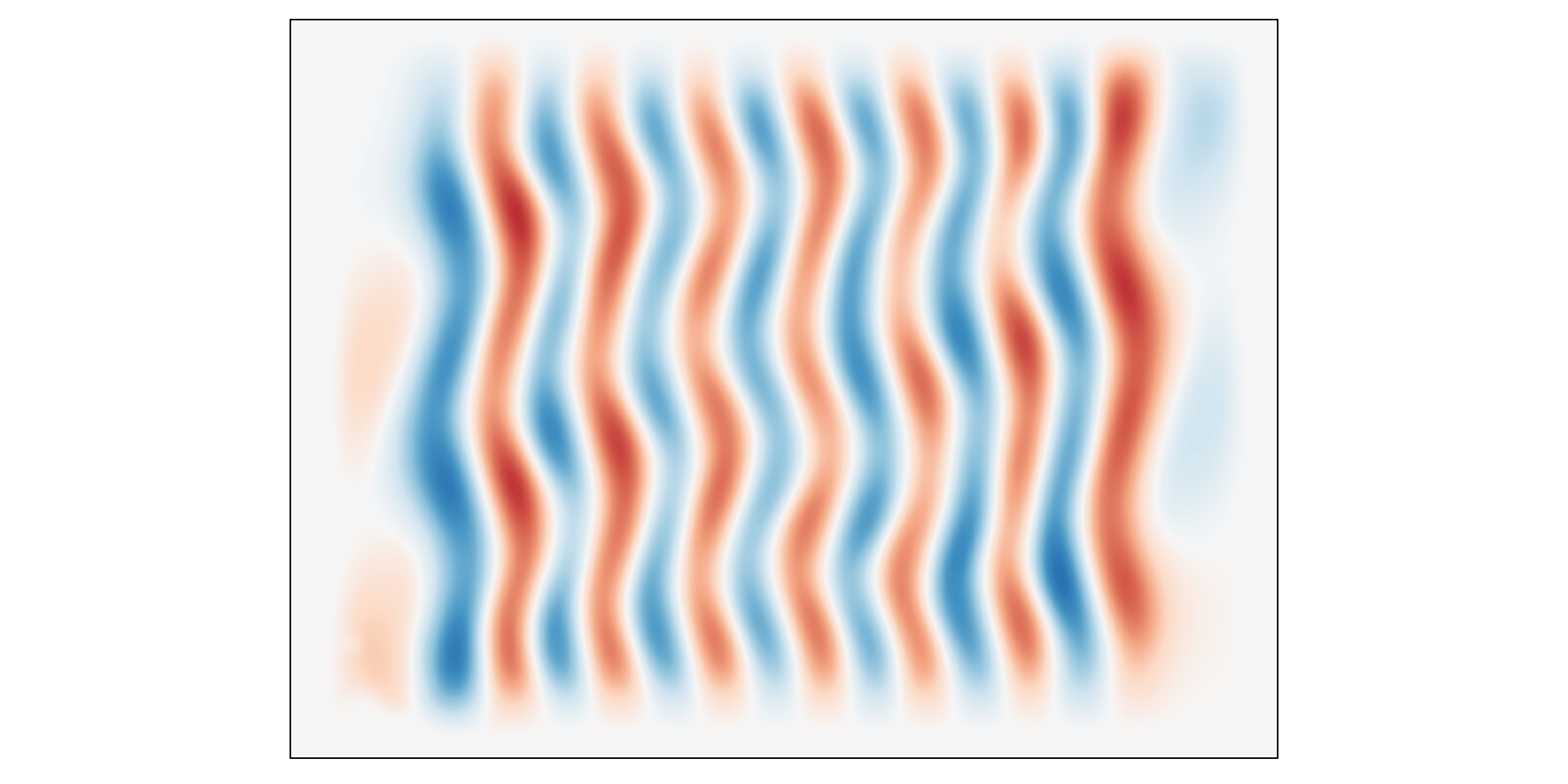} & \includegraphics[width=0.4\linewidth]{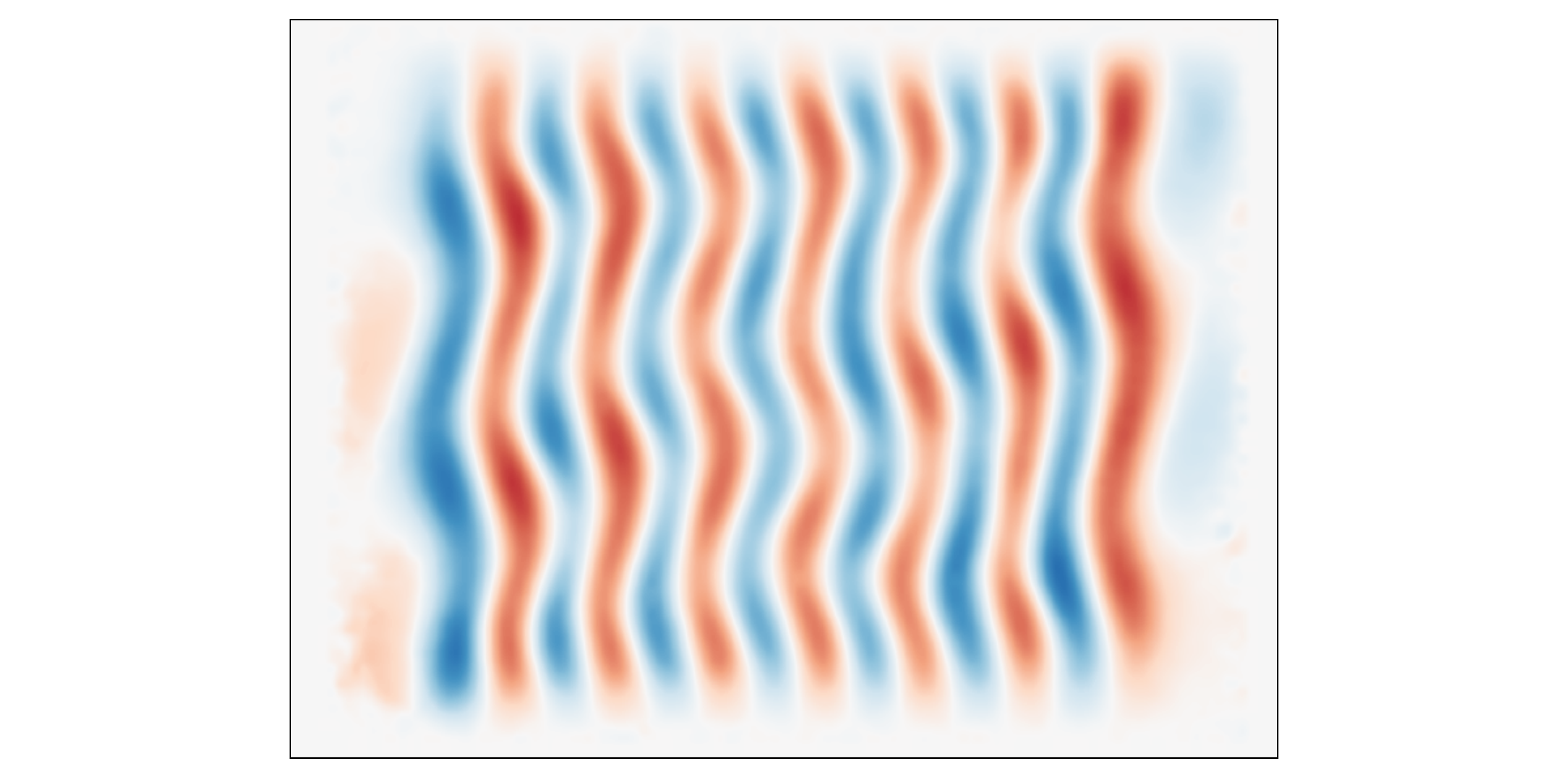} \\
\hline
\rotatebox[origin=l]{90}{\textbf{\ Cylinder flow}\ } & \rotatebox[origin=l]{90}{Rey. no.: $\approx 100$\ } & \includegraphics[width=0.4\linewidth]{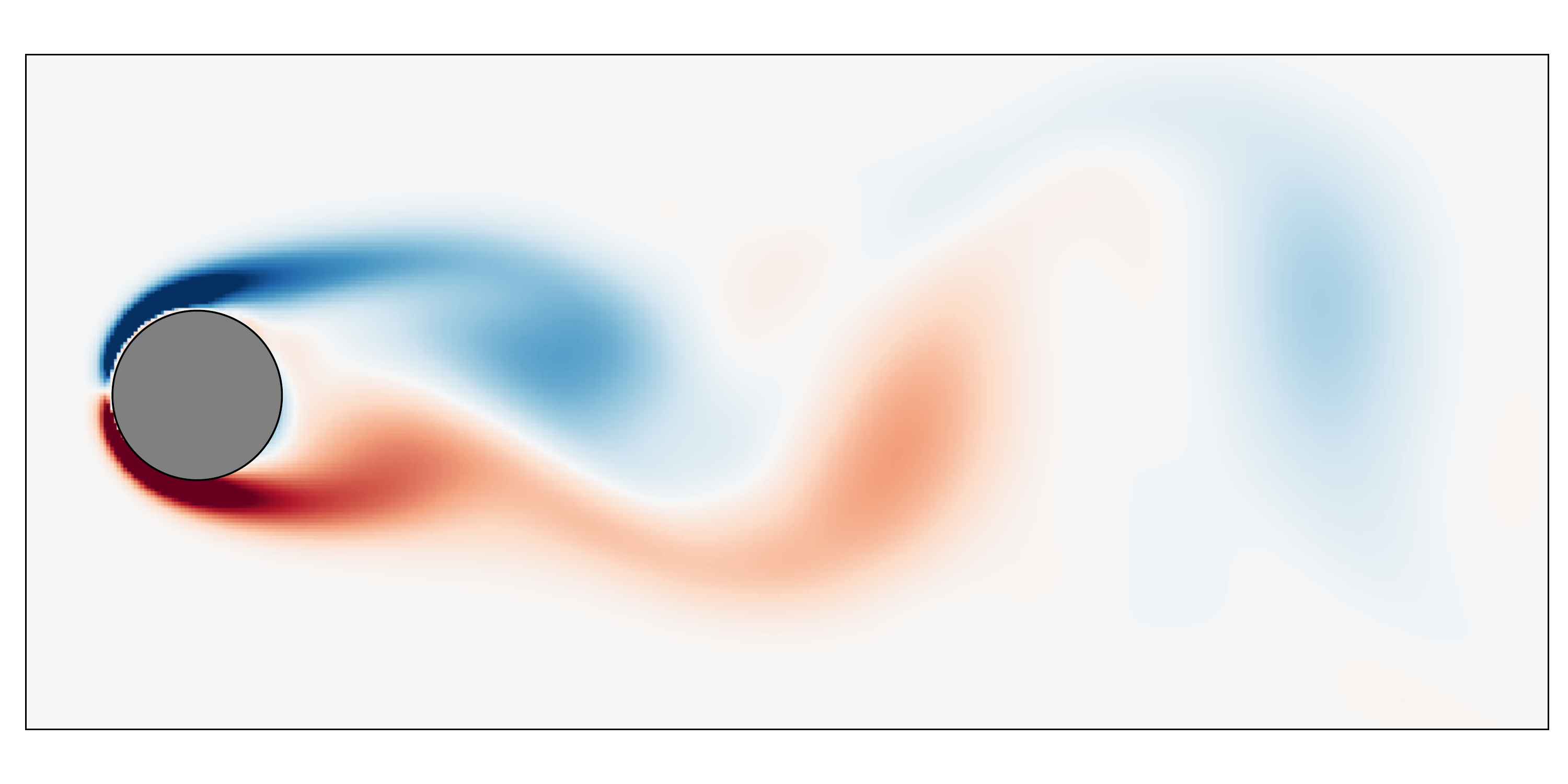} & \includegraphics[width=0.4\linewidth]{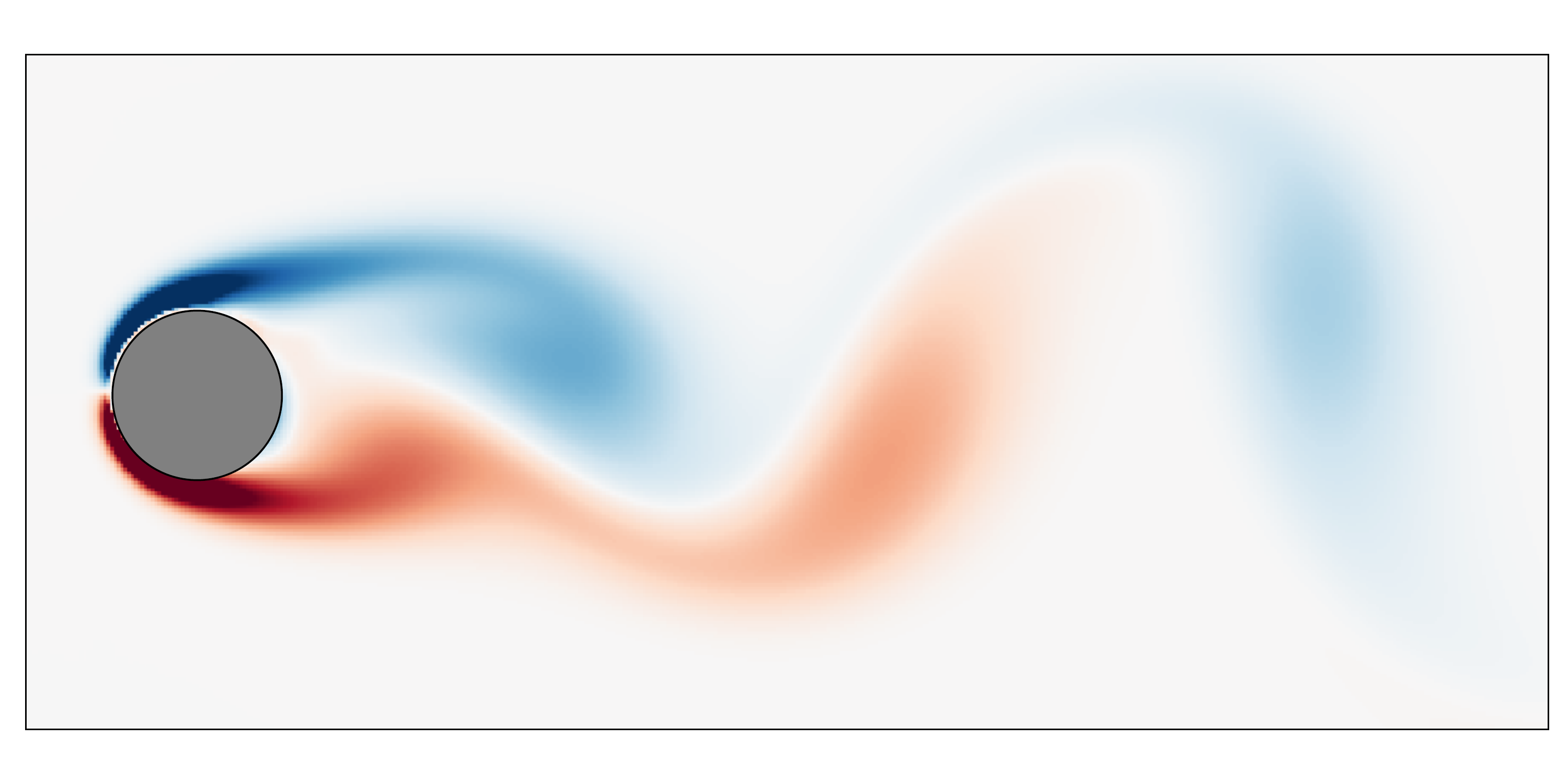}  \\
\hline
\rotatebox[origin=l]{90}{\textbf{Flow over cavity\ }}     & \rotatebox[origin=l]{90}{Rey. no.: $\approx 7500$\ } & \includegraphics[width=0.4\linewidth]{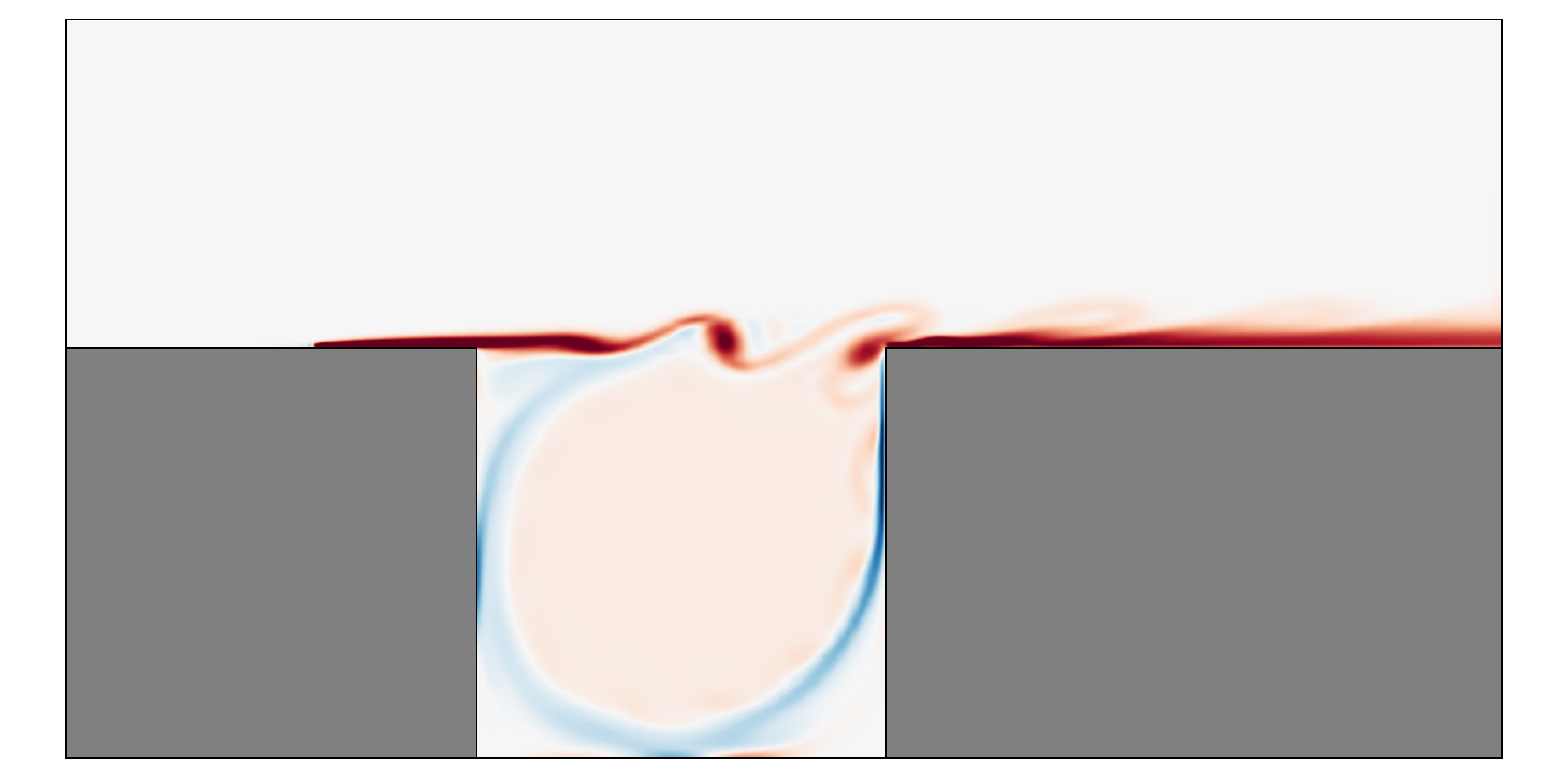} & \includegraphics[width=0.4\linewidth]{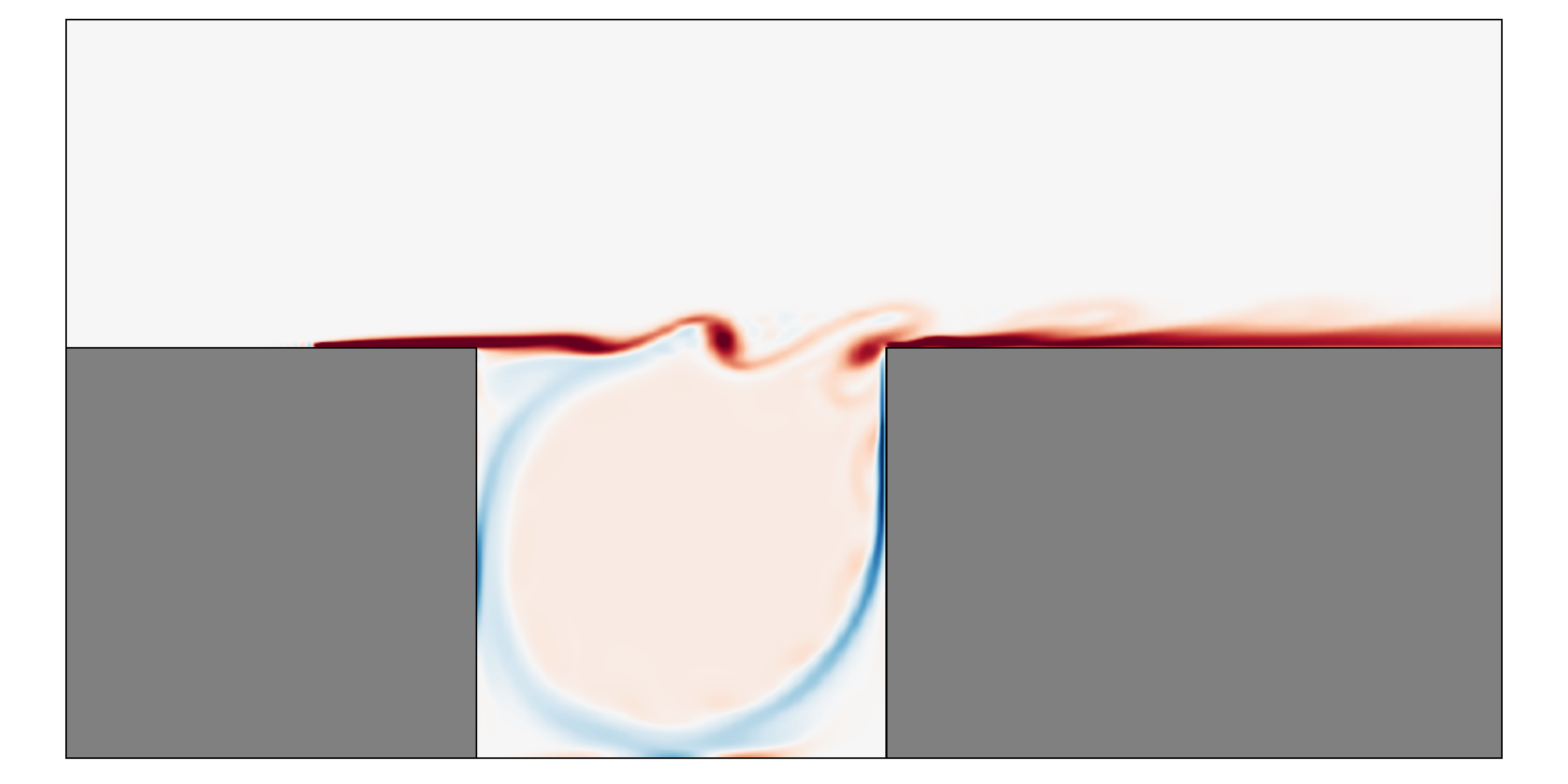} \\
\hline
\rotatebox[origin=l]{90}{\textbf{\ \ \ \ Video frame\ }}     & \rotatebox[origin=l]{90}{\ \ \ \ --} & \includegraphics[width=0.4\linewidth]{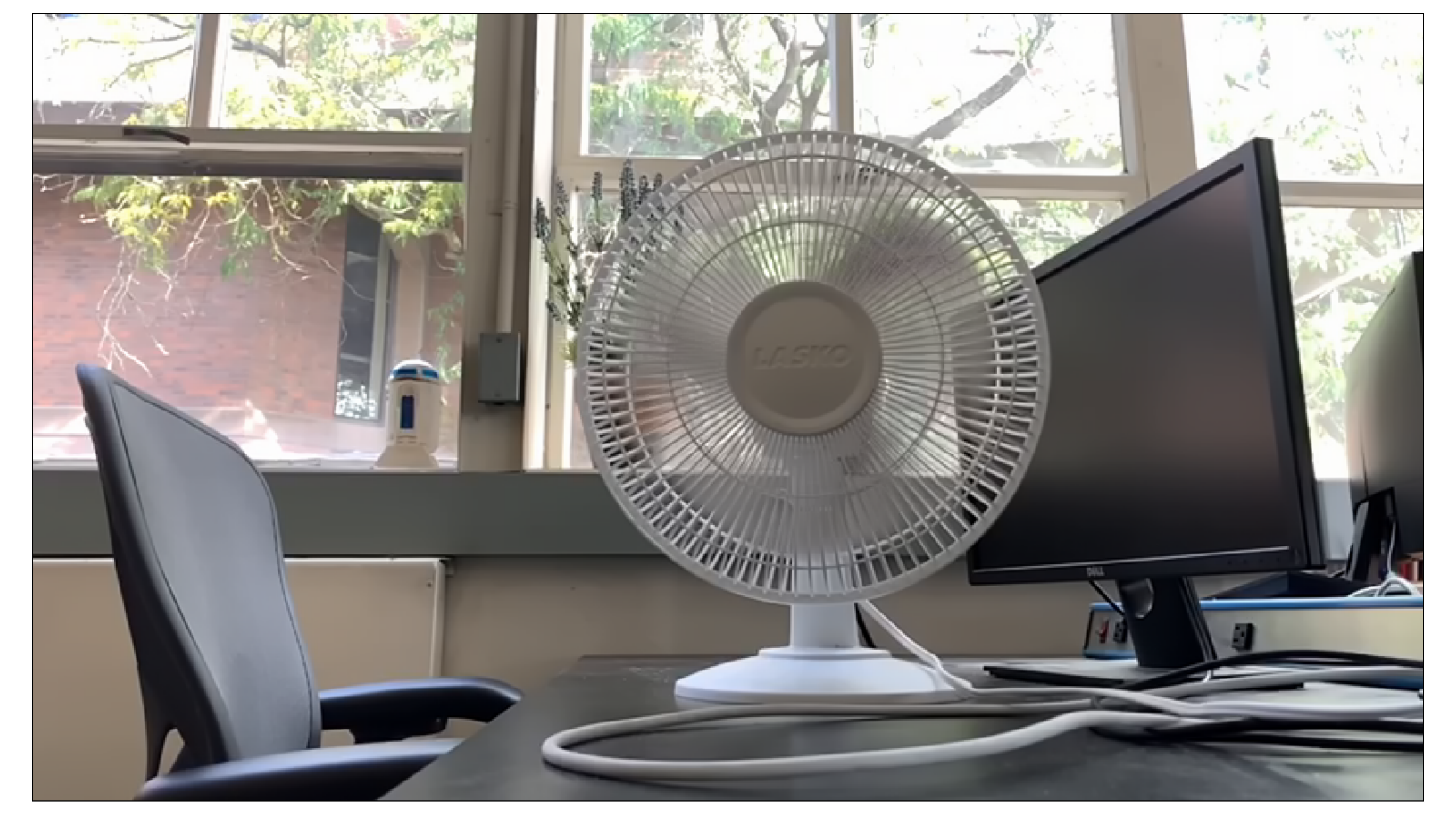} & \includegraphics[width=0.4\linewidth]{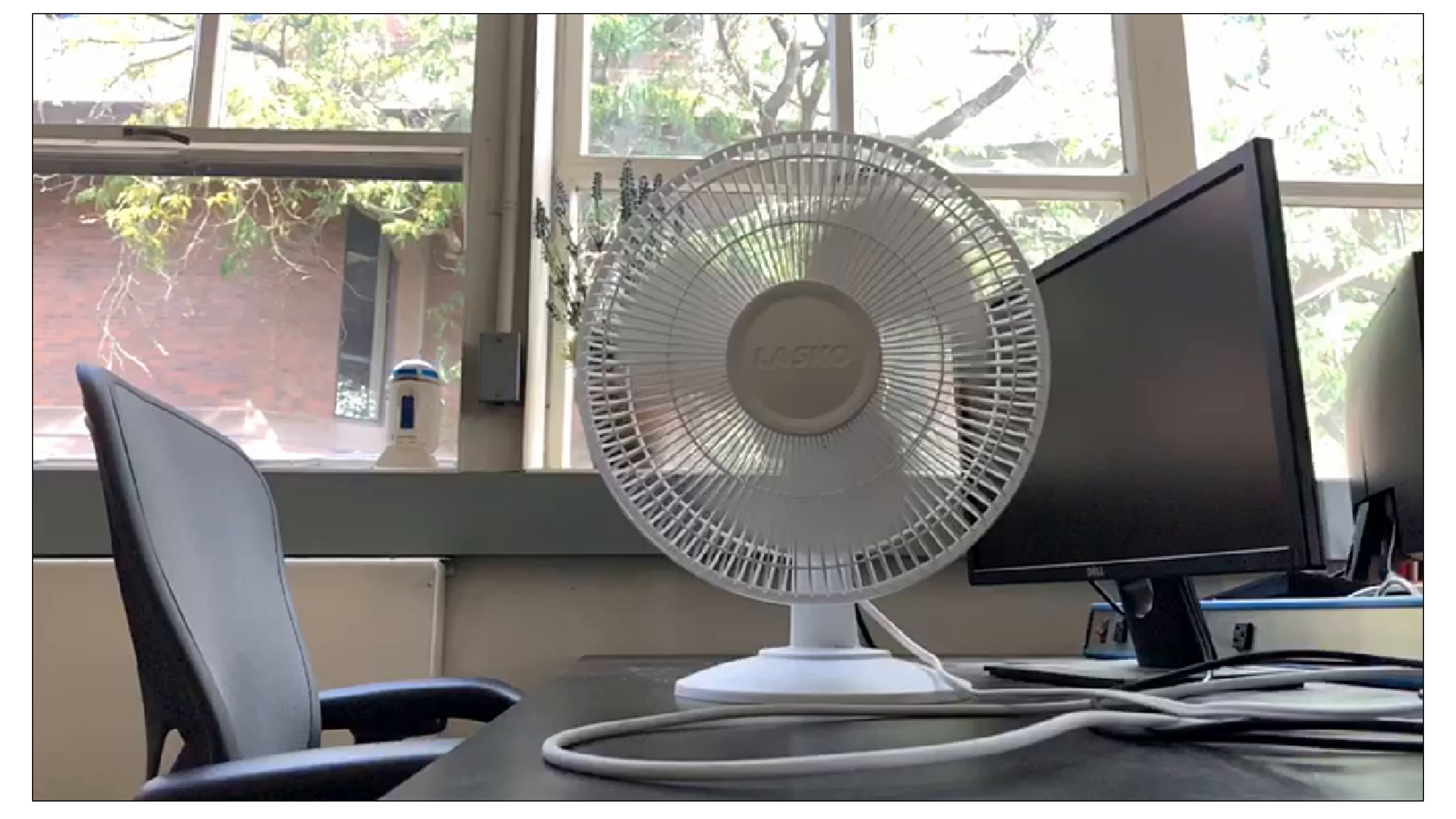} \\
\hline
\end{tabular}
\caption{The last frame as predicted by PCA in conjunction with the Fourier-based algorithm of fluid flows and video frame prediction. For a video that shows the performance, visit \emph{https://www.youtube.com/watch?v=trbXYMqi2Tw}.}
\label{table:fluid_flows}
\end{table}

Table \ref{table:fluid_flows} shows the performance of the Fourier based algorithm graphically, specifically it shows the last predicted frame when predicting a Kolmogorov 2D flow, flow around a cylinder, flow over a cavity and a video of a fan. The data from the Kolmogorov 2D flow was taken from the experiments conducted in ~\cite{tithof2017bifurcations}. Note that the Kolmogorov 2D flow and the data for the video frame prediction task constitute real measurements and therefore exhibit a considerable amount of noise, whereas the cylinder and cavity flow data stems from simulation.\\

\begin{table}[h!]
\begin{center}
\begin{tabular}{|l|l|l|l|l|l|}
\hline
                     & \textbf{Algorithm} & \multicolumn{4}{l|}{\textbf{Performance}} \\ \hline \hline
\textbf{Kolmogorov 2D}& \textbf{Fourier}   & $\mathbf{1.1}\%$ & $\mathbf{1.4\%}$ & $\mathbf{1.57\%}$ &  $\mathbf{1.69\%}$ \\ \cline{2-6}
Reynolds no. $\approx 10$& DMD   & $175\%$ & $171\%$ & $169\%$ & $165\%$\\ \cline{2-6}
\multirow{2}{*}{}    & FB DMD    & \emph{inf} & \emph{inf}  & \emph{inf} & \emph{inf} \\ \hline \hline
\textbf{Cylinder flow}& \textbf{Fourier}   & $\mathbf{0\%}$ & $\mathbf{0\%}$ & $\mathbf{0\%}$ & $\mathbf{0.1\%}$ \\ \cline{2-6}
Reynolds no. 100     & DMD       &  $4.44\%$  & $4.91\%$  & $4.71\%$  & $4.69\%$   \\ \cline{2-6}
\multirow{2}{*}{}    & FB DMD    &  $4.22\%$ & $4.75\%$ & $4.68\%$  &  $4.56\%$ \\ \hline \hline
\textbf{Flow over cavity}& \textbf{Fourier}   &  $\mathbf{0.9\%}$  &  $\mathbf{1\%}$  &  $\mathbf{1\%}$  &  $\mathbf{1\%}$  \\ \cline{2-6} 
Reynolds no. 7500    & DMD       &  $69\%$  & $66.7\%$ & $64.9\%$ & $63.8\%$\\ \cline{2-6} 
\multirow{2}{*}{}    & FB DMD    & $63\%$  & $70.6\%$ & $77.2\%$ & $84\%$  \\ \hline \hline
\textbf{Video frame} & Fourier   & $12.3\%$  & $17.2\%$  & $22.8\%$ &  $27.3\%$ \\ \cline{2-6} 
\multirow{3}{*}{}    & \textbf{DMD}       &  $\mathbf{7\%}$ & $\mathbf{11\%}$   & $\mathbf{16\%}$ & $\mathbf{22\%}$  \\ \cline{2-6} 
                     & FB DMD    & $194\%$ & $196\%$ & $196\%$ & $195\%$ \\ \hline \hline
Prediction Horizon  & &   25\%   & 50\%   & 75\%  & 100\% \\ \hline
\end{tabular}
\end{center}
\caption{The test set was split into four consecutive subsets of equal length. For every quantile, the relative cumulative error is reported for the Fourier-based algorithm, DMD and forward-backward DMD.}
\label{table:res_fluid_flows}
\end{table}

The test set was split into four equal length subsets and Table \ref{table:res_fluid_flows} shows the performance as measured by the relative cumulative error on each split. For each task, a parameter sweep over the number of frequencies was performed and the best performing model was selected. Note that the performance was measured in PCA-space, i.e. in predicting PCA modes. The results suggest that the Fourier-based algorithm significantly outperforms DMD and FB-DMD on three out of four tasks. The reason why DMD outperforms the Fourier-based algorithm on the video frame prediction task is simple: one of the PCA modes does not fit the inductive bias of the algorithm, i.e. it does not oscillate but it increases over time, therefore impeding the performance of the algorithm.

\subsection{Process noise and phase drifts}

The question arises whether process noise as opposed to measurement noise can be modeled with the algorithms introduced here. Human gait is known to be a oscillator that however exhibits a considerable amount of process noise~\citep{zielinska1996coupled}. We model process noise as a phase drift and slightly modify the optimization objective of the Fourier algorithm:
$$L(A, \Vec{\omega}, \phi_t, t) =  ||x_t - A\Omega(\Vec{\omega} t + \hat{\phi_t})||_2^2 + |\hat{\phi_t} - \hat{\phi}_{t-1}|$$

\begin{figure}
    \centering
    \includegraphics[width=0.75\linewidth]{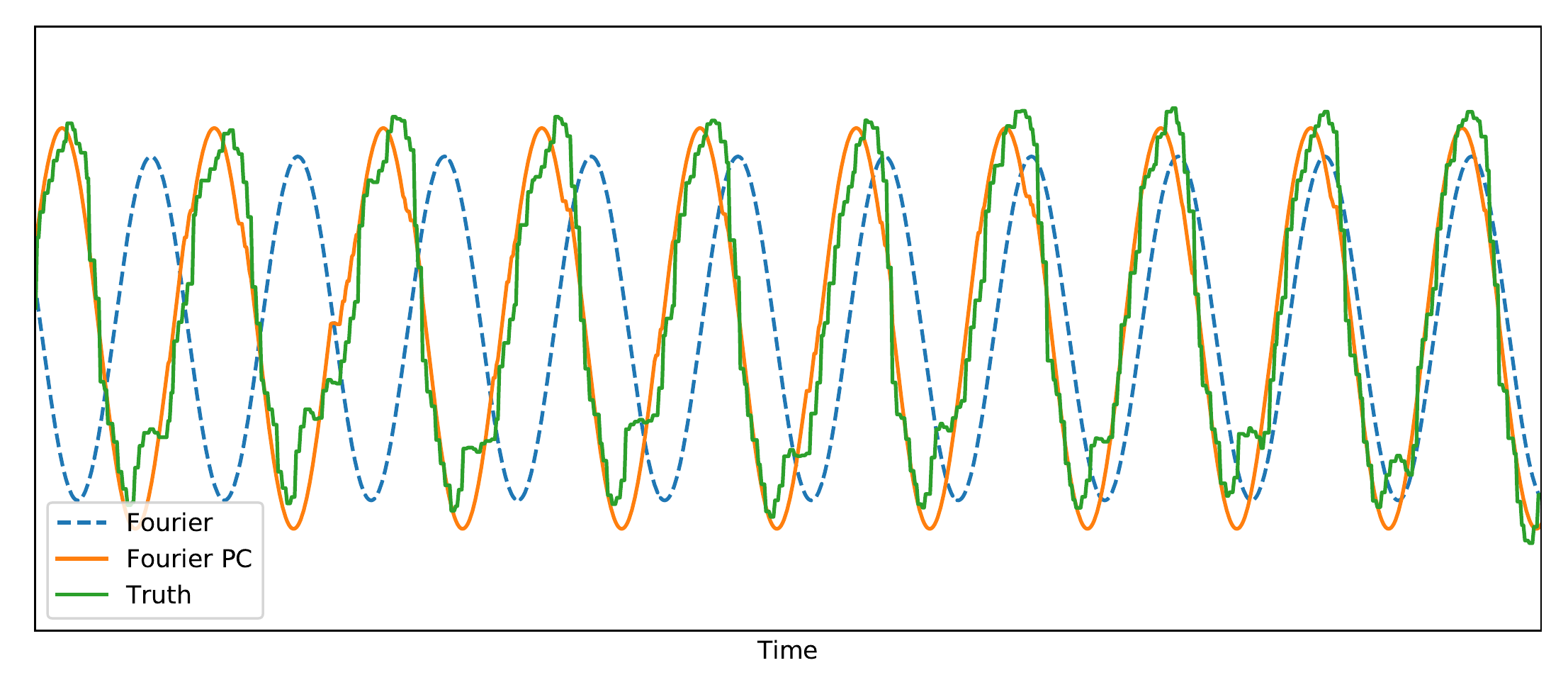}
    \caption{LocationZ recordings from subject 28. We visually compare the Fourier algorithm and its phase-corrected sibling (Fourier PC) to the true readings.}
    \label{fig:gait_pc}
\end{figure}
We test the modified algorithm on a data set containing accelerometer readings of a mobile phone located in test subjects pockets (subject 28, right pocket)~\citep{vajdi2019human} and compare to the Fourier algorithm. Figure \ref{fig:gait_pc} shows the first portion of the predictions of the Fourier algorithm alongside the phase-corrected modification (Fourier PC). Thus, small modifications to the algorithms introduced in this work allow to model process noise.

\section{Conclusion}

In this work, techniques to fit linear and nonlinear oscillators to data are introduced. When modeling quasi-periodic systems, the optimization objectives for both problems are highly non-convex and are therefore not amenable to be solved by gradient descent. The key insight to solving both optimization problems is to express the objectives in frequency domain where solutions can be found in a computationally efficient and scalable manner thanks to the computational properties of the Fast Fourier Transform. Because the optimization objectives are global, the algorithms introduced in this work allow for long-term forecasting while mitigating compounding errors. In the following, advantages and disadvantages of the algorithms are discussed.

\subsection{Disadvantages}
\begin{description}
    \item[Strong inductive bias:] The algorithms introduced here exhibit a strong inductive bias. The algorithms will not work well for data that stems from systems which are not \emph{quasi-periodic}. However, we have shown empirically that many real life phenomena such as fluid flows, space weather and energy consumption seem to fulfill this assumption.
    \item[Unknown phase problem:] This problem was alluded to earlier. Because analytical solutions for parameters $\Theta$ when $f$ is non-linear are usually unobtainable, swings in either direction from the `error plateau' can be indicative of a correct choice of frequency.
    \item[Conditional Independence of predictions:] Given model parameters, pairs of predicted values are conditionally independent, i.e. $p(x_{t_1}, x_{t_2} | \Theta, \Vec{\omega}) = p(x_{t_1} | \Theta, \Vec{\omega}) p(x_{t_2} | \Theta, \Vec{\omega})$. While this property is beneficial for \emph{compounding errors} and \emph{time stepping} (see below) this property can also have adverse effects. Consider the scenario where the goal is to perform a one-step ahead prediction for $t+1$ and all previous data points up to $t$ are known. If the model error for the prediction of $x_t$ is big, the prediction error for $x_{t+1}$ will most likely also be high. Note that in this scenario, we assume we know $x_t$ but even though, we know the true value of $x_t$, incorporating this knowledge, i.e. the conditional dependence on $x_{t+1}$, into the model is not trivial.
\end{description}

\subsection{Advantages}
\begin{description}
    \item[Strong inductive bias:] On the flip side, if the data is known to stem from a \emph{quasi-periodic} system, the algorithms introduced here offer a tight fit to the data. The resulting predictions are guaranteed to be quasi-periodic by construction therefore removing potential possibilities of failure like e.g. predictions exploding to plus/minus infinity.
    \item[Time stepping:] Past values are not required to make future state predictions when model parameters are known. This entails that time consuming time stepping methods are not required. This is especially useful if the data is high dimensional. This is closely related to the fact that predictions are conditionally independent.
    \item[Compounding errors:] Again, because a global optimization criterion (as opposed to one step-ahead predictions) is solved and because predictions are conditionally independent, errors do not compound. However, because of uncertainty caused by noise, errors will still accumulate over time but in a provably linear manner.
    \item[Noise resilience:] Because of the strong inductive bias, the algorithms introduced in this work exhibit very strong resiliency towards noise. As an anecdote, when Jaynes first published his results on \emph{Bayesian Spectral Analysis}~\citep{jaynes1987bayesian}, because he was able to resolve frequencies with much higher accuracy than Nyquist's theorem would suggest, his work was compared to `squaring a circle' and initially dismissed.
\end{description}

\section{Future Work}

In the opinion of the authors, the algorithms presented in this work create new and interesting potential future research paths. The following outlines some ideas for future work. %In the following, some ideas for future work are laid out.

\begin{description}
    \item[Input/output systems:] In its current inception, the algorithms introduced in this work assume the underlying systems to be autonomous, i.e. systems that are not driven by inputs. This severely limits the applicability of the algorithms in more general contexts. 
    Koopman theory and DMD have recently been extended to include externally forced systems~\citep{Proctor2016siads,Proctor2017siads}.
    % A class of algorithms can be conceived that make use of the central idea of Koopman theory, namely that nonlinear systems can be modeled as linear systems in conjunction with a time-invariant nonlinear function. 
    This idea could, in principle, be used to create new recurrent Neural Network (RNN) architectures that abandon nonlinearities in between time steps for linear systems in conjunction with a time-invariant nonlinearity. It may also lead to improved performance and could potentially overcome the issues that arise when training current RNNs, like e.g. exploding or vanishing gradients through time.
    \item[Interpretable flows:] When modeling the fluid flows little regard to the interpretability of the resulting model was given. Even though the flow over the cavity could, in principle, be modeled using a single frequency given an appropriate nonlinearity that is amenable to rotations, so far, the dimensionality was reduced using Principal Component Analysis and forecasting was performed using the Fourier-based algorithm. Employing the Koopman algorithm in conjunction with a nonlinear decoder that can handle rotations could reduce the number of required frequencies and result in higher noise robustness and interpretability.
    \item[Non-parametric Fourier/Koopman:] In this work, we showed a direct link between time series prediction and Bayesian Spectral Analysis (BSA). However, the insights from BSA are, so far, merely used to derive error bounds. These error bounds can, in principle, be used to influence predictions. One could conceive a Bayesian probabilistic model that marginalizes the frequencies out, i.e. by performing predictions by computing the following quantity: $\int p(x_{T+h} | \Vec{\omega}) p(\Vec{\omega}|x_{1:T}) d\Vec{\omega}$. Note that Bayesian Spectral Analysis provides ways to calculate $p(\Vec{\omega}|x_{1:T})$.
    \item[Local and global modeling:] Consider the following scenario: The goal is to predict a signal with seasonal patterns given multiple years of training data. Imagine also that the current year is anomalous which entails that short-term predictions will be impeded. In its current inception, because global optima are sought, the information that the system is currently in an anomalous state is hard to incorporate. Ideally, in addition to one of the global models introduced in this work, there is also a local model and with increasing prediction horizon, more weight should be given to the global over the local model. Optimally weighing local and global models seem an interesting and important future research path.
\end{description}

\section*{Acknowledgements}

The authors acknowledge support from the Defense Advanced Research Projects Agency (DARPA PA-18-01-FP-125) and the Army Research Office (ARO W911NF-17-1-0306). We would like to thank Charles Delahunt for feedback on the manuscript, and Jeffrey Tithof, Jared Callaham, Kadierdan Kaheman, Benjamin Herrmann and Isabel Scherl for providing data sets.

\bibliography{references}

\begin{thebibliography}{66}
\providecommand{\natexlab}[1]{#1}
\providecommand{\url}[1]{\texttt{#1}}
\expandafter\ifx\csname urlstyle\endcsname\relax
  \providecommand{\doi}[1]{doi: #1}\else
  \providecommand{\doi}{doi: \begingroup \urlstyle{rm}\Url}\fi

\bibitem[Armstrong(1985)]{armstrong1985long}
J.~S. Armstrong.
\newblock \emph{Long-range forecasting}.
\newblock Wiley New York ETC., 1985.

\bibitem[Askham and Kutz(2018)]{askham2018variable}
T.~Askham and J.~N. Kutz.
\newblock Variable projection methods for an optimized dynamic mode
  decomposition.
\newblock \emph{SIAM Journal on Applied Dynamical Systems}, 17\penalty0
  (1):\penalty0 380--416, 2018.

\bibitem[Birkhoff(1931)]{birkhoff1931proof}
G.~D. Birkhoff.
\newblock Proof of the ergodic theorem.
\newblock \emph{Proceedings of the National Academy of Sciences}, 17\penalty0
  (12):\penalty0 656--660, 1931.

\bibitem[Birkhoff and Koopman(1932)]{birkhoff1932recent}
G.~D. Birkhoff and B.~Koopman.
\newblock Recent contributions to the ergodic theory.
\newblock \emph{Proceedings of the National Academy of Sciences of the United
  States of America}, 18\penalty0 (3):\penalty0 279, 1932.

\bibitem[Box et~al.(2015)Box, Jenkins, Reinsel, and Ljung]{box2015time}
G.~E. Box, G.~M. Jenkins, G.~C. Reinsel, and G.~M. Ljung.
\newblock \emph{Time series analysis: forecasting and control}.
\newblock John Wiley \& Sons, 2015.

\bibitem[Bretthorst(2013)]{bretthorst2013bayesian}
G.~L. Bretthorst.
\newblock \emph{Bayesian spectrum analysis and parameter estimation},
  volume~48.
\newblock Springer Science \& Business Media, 2013.

\bibitem[Brunton et~al.(2016{\natexlab{a}})Brunton, Johnson, Ojemann, and
  Kutz]{brunton2016extracting}
B.~W. Brunton, L.~A. Johnson, J.~G. Ojemann, and J.~N. Kutz.
\newblock Extracting spatial--temporal coherent patterns in large-scale neural
  recordings using dynamic mode decomposition.
\newblock \emph{Journal of neuroscience methods}, 258:\penalty0 1--15,
  2016{\natexlab{a}}.

\bibitem[Brunton et~al.(2016{\natexlab{b}})Brunton, Brunton, Proctor, and
  Kutz]{Brunton2016plosone}
S.~L. Brunton, B.~W. Brunton, J.~L. Proctor, and J.~N. Kutz.
\newblock Koopman invariant subspaces and finite linear representations of
  nonlinear dynamical systems for control.
\newblock \emph{PLoS ONE}, 11\penalty0 (2):\penalty0 e0150171,
  2016{\natexlab{b}}.

\bibitem[Brunton et~al.(2017)Brunton, Brunton, Proctor, Kaiser, and
  Kutz]{brunton2017chaos}
S.~L. Brunton, B.~W. Brunton, J.~L. Proctor, E.~Kaiser, and J.~N. Kutz.
\newblock Chaos as an intermittently forced linear system.
\newblock \emph{Nature Communications}, 8\penalty0 (1):\penalty0 1--9, 2017.

\bibitem[Budi{\v{s}}i{\'c} et~al.(2012)Budi{\v{s}}i{\'c}, Mohr, and
  Mezi{\'c}]{budivsic2012applied}
M.~Budi{\v{s}}i{\'c}, R.~Mohr, and I.~Mezi{\'c}.
\newblock Applied {K}oopmanism.
\newblock \emph{Chaos: An Interdisciplinary Journal of Nonlinear Science},
  22\penalty0 (4):\penalty0 047510, 2012.

\bibitem[Champion et~al.(2019)Champion, Lusch, Kutz, and
  Brunton]{champion2019data}
K.~Champion, B.~Lusch, J.~N. Kutz, and S.~L. Brunton.
\newblock Data-driven discovery of coordinates and governing equations.
\newblock \emph{Proceedings of the National Academy of Sciences}, 116\penalty0
  (45):\penalty0 22445--22451, 2019.

\bibitem[Chan et~al.(2012)Chan, Tsui, Wu, Hou, Wu, and Wu]{chan2012load}
S.-C. Chan, K.~M. Tsui, H.~Wu, Y.~Hou, Y.-C. Wu, and F.~F. Wu.
\newblock Load/price forecasting and managing demand response for smart grids:
  Methodologies and challenges.
\newblock \emph{IEEE signal processing magazine}, 29\penalty0 (5):\penalty0
  68--85, 2012.

\bibitem[Chen et~al.(2012)Chen, Tu, and Rowley]{chen2012variants}
K.~K. Chen, J.~H. Tu, and C.~W. Rowley.
\newblock Variants of dynamic mode decomposition: boundary condition,
  {K}oopman, and {F}ourier analyses.
\newblock \emph{Journal of nonlinear science}, 22\penalty0 (6):\penalty0
  887--915, 2012.

\bibitem[Chung et~al.(2014)Chung, Gulcehre, Cho, and
  Bengio]{chung2014empirical}
J.~Chung, C.~Gulcehre, K.~Cho, and Y.~Bengio.
\newblock Empirical evaluation of gated recurrent neural networks on sequence
  modeling.
\newblock \emph{arXiv preprint arXiv:1412.3555}, 2014.

\bibitem[Cooley and Tukey(1965)]{cooley1965algorithm}
J.~W. Cooley and J.~W. Tukey.
\newblock An algorithm for the machine calculation of complex {F}ourier series.
\newblock \emph{Mathematics of computation}, 19\penalty0 (90):\penalty0
  297--301, 1965.

\bibitem[Dawson et~al.(2016)Dawson, Hemati, Williams, and
  Rowley]{dawson2016characterizing}
S.~T. Dawson, M.~S. Hemati, M.~O. Williams, and C.~W. Rowley.
\newblock Characterizing and correcting for the effect of sensor noise in the
  dynamic mode decomposition.
\newblock \emph{Experiments in Fluids}, 57\penalty0 (3):\penalty0 42, 2016.

\bibitem[Erichson et~al.(2019{\natexlab{a}})Erichson, Brunton, and
  Kutz]{erichson2015compressed}
N.~B. Erichson, S.~L. Brunton, and J.~N. Kutz.
\newblock Compressed dynamic mode decomposition for real-time object detection.
\newblock \emph{Journal of Real-Time Image Processing}, 16\penalty0
  (5):\penalty0 1479--1492, 2019{\natexlab{a}}.

\bibitem[Erichson et~al.(2019{\natexlab{b}})Erichson, Muehlebach, and
  Mahoney]{erichson2019physics}
N.~B. Erichson, M.~Muehlebach, and M.~W. Mahoney.
\newblock Physics-informed autoencoders for {L}yapunov-stable fluid flow
  prediction.
\newblock \emph{arXiv preprint arXiv:1905.10866}, 2019{\natexlab{b}}.

\bibitem[Fourier(1822)]{fourier1822theorie}
J.~Fourier.
\newblock \emph{Theorie analytique de la chaleur, par M. Fourier}.
\newblock Chez Firmin Didot, p{\`e}re et fils, 1822.

\bibitem[Gashler and Ashmore(2016)]{gashler2016modeling}
M.~S. Gashler and S.~C. Ashmore.
\newblock Modeling time series data with deep {F}ourier neural networks.
\newblock \emph{Neurocomputing}, 188:\penalty0 3--11, 2016.

\bibitem[Hamilton(1994)]{hamilton1994time}
J.~D. Hamilton.
\newblock \emph{Time series analysis}, volume~2.
\newblock Princeton New Jersey, 1994.

\bibitem[Hemati et~al.(2017)Hemati, Rowley, Deem, and
  Cattafesta]{hemati2017biasing}
M.~S. Hemati, C.~W. Rowley, E.~A. Deem, and L.~N. Cattafesta.
\newblock De-biasing the dynamic mode decomposition for applied {K}oopman
  spectral analysis of noisy datasets.
\newblock \emph{Theoretical and Computational Fluid Dynamics}, 31\penalty0
  (4):\penalty0 349--368, 2017.

\bibitem[Hochreiter and Schmidhuber(1997)]{hochreiter1997long}
S.~Hochreiter and J.~Schmidhuber.
\newblock Long short-term memory.
\newblock \emph{Neural computation}, 9\penalty0 (8):\penalty0 1735--1780, 1997.

\bibitem[Hyndman et~al.(2007)Hyndman, Khandakar, et~al.]{hyndman2007automatic}
R.~J. Hyndman, Y.~Khandakar, et~al.
\newblock \emph{Automatic time series for forecasting: the forecast package for
  R}.
\newblock Number 6/07. Monash University, Department of Econometrics and
  Business Statistics~…, 2007.

\bibitem[Jaeger(2007)]{jaeger2007echo}
H.~Jaeger.
\newblock Echo state network.
\newblock \emph{scholarpedia}, 2\penalty0 (9):\penalty0 2330, 2007.

\bibitem[Jaynes(1987)]{jaynes1987bayesian}
E.~Jaynes.
\newblock Bayesian spectrum and chirp analysis.
\newblock In \emph{Maximum-Entropy and Bayesian Spectral Analysis and
  Estimation Problems}, pages 1--37. Springer, 1987.

\bibitem[Jensen and Pinson(2017)]{jensen2017re}
T.~V. Jensen and P.~Pinson.
\newblock Re-europe, a large-scale dataset for modeling a highly renewable
  european electricity system.
\newblock \emph{Scientific data}, 4:\penalty0 170175, 2017.

\bibitem[Jiang and Adeli(2005)]{jiang2005dynamic}
X.~Jiang and H.~Adeli.
\newblock Dynamic wavelet neural network model for traffic flow forecasting.
\newblock \emph{Journal of transportation engineering}, 131\penalty0
  (10):\penalty0 771--779, 2005.

\bibitem[Kamb et~al.(2018)Kamb, Kaiser, Brunton, and Kutz]{kamb2018time}
M.~Kamb, E.~Kaiser, S.~L. Brunton, and J.~N. Kutz.
\newblock Time-delay observables for {K}oopman: Theory and applications.
\newblock \emph{arXiv preprint arXiv:1810.01479}, 2018.

\bibitem[Koopman(1931)]{koopman1931hamiltonian}
B.~O. Koopman.
\newblock Hamiltonian systems and transformation in {H}ilbert space.
\newblock \emph{Proceedings of the National Academy of Sciences of the United
  States of America}, 17\penalty0 (5):\penalty0 315, 1931.

\bibitem[Koopman and Neumann(1932)]{koopman1932dynamical}
B.~O. Koopman and J.~v. Neumann.
\newblock Dynamical systems of continuous spectra.
\newblock \emph{Proceedings of the National Academy of Sciences of the United
  States of America}, 18\penalty0 (3):\penalty0 255, 1932.

\bibitem[Koutnik et~al.(2014)Koutnik, Greff, Gomez, and
  Schmidhuber]{koutnik2014clockwork}
J.~Koutnik, K.~Greff, F.~Gomez, and J.~Schmidhuber.
\newblock A clockwork {rnn}.
\newblock \emph{arXiv preprint arXiv:1402.3511}, 2014.

\bibitem[Kutz et~al.(2016)Kutz, Brunton, Brunton, and Proctor]{kutz2016dynamic}
J.~N. Kutz, S.~L. Brunton, B.~W. Brunton, and J.~L. Proctor.
\newblock \emph{Dynamic mode decomposition: data-driven modeling of complex
  systems}.
\newblock SIAM, 2016.

\bibitem[Lan and Mezi{\'c}(2013)]{lan2013linearization}
Y.~Lan and I.~Mezi{\'c}.
\newblock Linearization in the large of nonlinear systems and {K}oopman
  operator spectrum.
\newblock \emph{Physica D: Nonlinear Phenomena}, 242\penalty0 (1):\penalty0
  42--53, 2013.

\bibitem[LeCun et~al.(2015)LeCun, Bengio, and Hinton]{lecun2015deep}
Y.~LeCun, Y.~Bengio, and G.~Hinton.
\newblock Deep learning.
\newblock \emph{Nature}, 521\penalty0 (7553):\penalty0 436, 2015.

\bibitem[Lusch et~al.(2018)Lusch, Kutz, and Brunton]{lusch2018deep}
B.~Lusch, J.~N. Kutz, and S.~L. Brunton.
\newblock Deep learning for universal linear embeddings of nonlinear dynamics.
\newblock \emph{Nature Communications}, 9\penalty0 (1):\penalty0 4950, 2018.

\bibitem[Lyon(2009)]{lyon2009discrete}
D.~A. Lyon.
\newblock The discrete {F}ourier transform, part 4: spectral leakage.
\newblock \emph{Journal of object technology}, 8\penalty0 (7), 2009.

\bibitem[Mezi{\'c}(2005)]{mezic2005spectral}
I.~Mezi{\'c}.
\newblock Spectral properties of dynamical systems, model reduction and
  decompositions.
\newblock \emph{Nonlinear Dynamics}, 41\penalty0 (1-3):\penalty0 309--325,
  2005.

\bibitem[Mezic(2013)]{Mezic2013arfm}
I.~Mezic.
\newblock Analysis of fluid flows via spectral properties of the {K}oopman
  operator.
\newblock \emph{Annual Review of Fluid Mechanics}, 45:\penalty0 357--378, 2013.

\bibitem[Mezi{\'c}(2015)]{mezic2015applications}
I.~Mezi{\'c}.
\newblock On applications of the spectral theory of the {K}oopman operator in
  dynamical systems and control theory.
\newblock In \emph{2015 54th IEEE Conference on Decision and Control (CDC)},
  pages 7034--7041. IEEE, 2015.

\bibitem[Moeck et~al.(2013)Moeck, Bourgouin, Durox, Schuller, and
  Candel]{moeck2013tomographic}
J.~P. Moeck, J.-F. Bourgouin, D.~Durox, T.~Schuller, and S.~Candel.
\newblock Tomographic reconstruction of heat release rate perturbations induced
  by helical modes in turbulent swirl flames.
\newblock \emph{Experiments in fluids}, 54\penalty0 (4):\penalty0 1498, 2013.

\bibitem[Moore(2015)]{moore2015ergodic}
C.~C. Moore.
\newblock Ergodic theorem, ergodic theory, and statistical mechanics.
\newblock \emph{Proceedings of the National Academy of Sciences}, 112\penalty0
  (7):\penalty0 1907--1911, 2015.

\bibitem[Neumann(1932{\natexlab{a}})]{neumann1932physical}
J.~v. Neumann.
\newblock Physical applications of the ergodic hypothesis.
\newblock \emph{Proceedings of the National Academy of Sciences of the United
  States of America}, 18\penalty0 (3):\penalty0 263, 1932{\natexlab{a}}.

\bibitem[Neumann(1932{\natexlab{b}})]{neumann1932proof}
J.~v. Neumann.
\newblock Proof of the quasi-ergodic hypothesis.
\newblock \emph{Proceedings of the National Academy of Sciences}, 18\penalty0
  (1):\penalty0 70--82, 1932{\natexlab{b}}.

\bibitem[Otto and Rowley(2019)]{otto2019linearly}
S.~E. Otto and C.~W. Rowley.
\newblock Linearly recurrent autoencoder networks for learning dynamics.
\newblock \emph{SIAM Journal on Applied Dynamical Systems}, 18\penalty0
  (1):\penalty0 558--593, 2019.

\bibitem[Pan and Duraisamy(2019{\natexlab{a}})]{pan2019physics}
S.~Pan and K.~Duraisamy.
\newblock Physics-informed probabilistic learning of linear embeddings of
  non-linear dynamics with guaranteed stability.
\newblock \emph{arXiv preprint arXiv:1906.03663}, 2019{\natexlab{a}}.

\bibitem[Pan and Duraisamy(2019{\natexlab{b}})]{pan2019structure}
S.~Pan and K.~Duraisamy.
\newblock On the structure of time-delay embedding in linear models of
  non-linear dynamical systems.
\newblock \emph{arXiv preprint arXiv:1902.05198}, 2019{\natexlab{b}}.

\bibitem[Pankratz(2009)]{pankratz2009forecasting}
A.~Pankratz.
\newblock \emph{Forecasting with univariate Box-Jenkins models: Concepts and
  cases}, volume 224.
\newblock John Wiley \& Sons, 2009.

\bibitem[Paquot et~al.(2012)Paquot, Duport, Smerieri, Dambre, Schrauwen,
  Haelterman, and Massar]{paquot2012optoelectronic}
Y.~Paquot, F.~Duport, A.~Smerieri, J.~Dambre, B.~Schrauwen, M.~Haelterman, and
  S.~Massar.
\newblock Optoelectronic reservoir computing.
\newblock \emph{Scientific reports}, 2:\penalty0 287, 2012.

\bibitem[Proctor and Eckhoff(2015)]{proctor2015discovering}
J.~L. Proctor and P.~A. Eckhoff.
\newblock Discovering dynamic patterns from infectious disease data using
  dynamic mode decomposition.
\newblock \emph{International health}, 7\penalty0 (2):\penalty0 139--145, 2015.

\bibitem[Proctor et~al.(2016)Proctor, Brunton, and Kutz]{Proctor2016siads}
J.~L. Proctor, S.~L. Brunton, and J.~N. Kutz.
\newblock Dynamic mode decomposition with control.
\newblock \emph{SIAM Journal on Applied Dynamical Systems}, 15\penalty0
  (1):\penalty0 142--161, 2016.

\bibitem[Proctor et~al.(2018)Proctor, Brunton, and Kutz]{Proctor2017siads}
J.~L. Proctor, S.~L. Brunton, and J.~N. Kutz.
\newblock Generalizing {K}oopman theory to allow for inputs and control.
\newblock \emph{SIAM Journal on Applied Dynamical Systems}, 17\penalty0
  (1):\penalty0 909--930, 2018.

\bibitem[Rowley et~al.(2009)Rowley, Mezi{\'c}, Bagheri, Schlatter, and
  Henningson]{rowley2009spectral}
C.~W. Rowley, I.~Mezi{\'c}, S.~Bagheri, P.~Schlatter, and D.~S. Henningson.
\newblock Spectral analysis of nonlinear flows.
\newblock \emph{Journal of fluid mechanics}, 641:\penalty0 115--127, 2009.

\bibitem[Sakamoto et~al.(1986)Sakamoto, Ishiguro, and
  Kitagawa]{sakamoto1986akaike}
Y.~Sakamoto, M.~Ishiguro, and G.~Kitagawa.
\newblock Akaike information criterion statistics.
\newblock \emph{Dordrecht, The Netherlands: D. Reidel}, 81, 1986.

\bibitem[Schmid(2010)]{schmid2010dynamic}
P.~J. Schmid.
\newblock Dynamic mode decomposition of numerical and experimental data.
\newblock \emph{Journal of fluid mechanics}, 656:\penalty0 5--28, 2010.

\bibitem[Scholkopf and Smola(2001)]{scholkopf2001learning}
B.~Scholkopf and A.~J. Smola.
\newblock \emph{Learning with kernels: support vector machines, regularization,
  optimization, and beyond}.
\newblock MIT press, 2001.

\bibitem[Silvescu(1999)]{silvescu1999fourier}
A.~Silvescu.
\newblock Fourier neural networks.
\newblock In \emph{IJCNN'99. International Joint Conference on Neural Networks.
  Proceedings}, volume~1, pages 488--491. IEEE, 1999.

\bibitem[Song et~al.(2013)Song, Alizard, Robinet, and
  Gloerfelt]{song2013global}
G.~Song, F.~Alizard, J.-C. Robinet, and X.~Gloerfelt.
\newblock Global and {K}oopman modes analysis of sound generation in mixing
  layers.
\newblock \emph{Physics of Fluids}, 25\penalty0 (12):\penalty0 124101, 2013.

\bibitem[Takeishi et~al.(2017)Takeishi, Kawahara, and
  Yairi]{takeishi2017learning}
N.~Takeishi, Y.~Kawahara, and T.~Yairi.
\newblock Learning {K}oopman invariant subspaces for dynamic mode
  decomposition.
\newblock In \emph{Advances in Neural Information Processing Systems}, pages
  1130--1140, 2017.

\bibitem[Tithof et~al.(2017)Tithof, Suri, Pallantla, Grigoriev, and
  Schatz]{tithof2017bifurcations}
J.~Tithof, B.~Suri, R.~K. Pallantla, R.~O. Grigoriev, and M.~F. Schatz.
\newblock Bifurcations in a quasi-two-dimensional kolmogorov-like flow.
\newblock \emph{Journal of Fluid Mechanics}, 828:\penalty0 837--866, 2017.

\bibitem[Tseng(2001)]{tseng2001convergence}
P.~Tseng.
\newblock Convergence of a block coordinate descent method for
  nondifferentiable minimization.
\newblock \emph{Journal of optimization theory and applications}, 109\penalty0
  (3):\penalty0 475--494, 2001.

\bibitem[Tu et~al.(2014)Tu, Rowley, Luchtenburg, Brunton, and Kutz]{Tu2014jcd}
J.~H. Tu, C.~W. Rowley, D.~M. Luchtenburg, S.~L. Brunton, and J.~N. Kutz.
\newblock On dynamic mode decomposition: theory and applications.
\newblock \emph{Journal of Computational Dynamics}, 1\penalty0 (2):\penalty0
  391--421, 2014.

\bibitem[Vajdi et~al.(2019)Vajdi, Zaghian, Farahmand, Rastegar, Maroofi, Jia,
  Pomplun, Haspel, and Bayat]{vajdi2019human}
A.~Vajdi, M.~R. Zaghian, S.~Farahmand, E.~Rastegar, K.~Maroofi, S.~Jia,
  M.~Pomplun, N.~Haspel, and A.~Bayat.
\newblock Human gait database for normal walk collected by smart phone
  accelerometer.
\newblock \emph{arXiv preprint arXiv:1905.03109}, 2019.

\bibitem[Wehmeyer and No{\'e}(2018)]{wehmeyer2018time}
C.~Wehmeyer and F.~No{\'e}.
\newblock Time-lagged autoencoders: Deep learning of slow collective variables
  for molecular kinetics.
\newblock \emph{The Journal of chemical physics}, 148\penalty0 (24):\penalty0
  241703, 2018.

\bibitem[Yeung et~al.(2019)Yeung, Kundu, and Hodas]{yeung2019learning}
E.~Yeung, S.~Kundu, and N.~Hodas.
\newblock Learning deep neural network representations for {K}oopman operators
  of nonlinear dynamical systems.
\newblock In \emph{2019 American Control Conference (ACC)}, pages 4832--4839.
  IEEE, 2019.

\bibitem[Zieli{\'n}ska(1996)]{zielinska1996coupled}
T.~Zieli{\'n}ska.
\newblock Coupled oscillators utilised as gait rhythm generators of a
  two-legged walking machine.
\newblock \emph{Biological Cybernetics}, 74\penalty0 (3):\penalty0 263--273,
  1996.

\end{thebibliography}
\end{document}